\documentclass[5p,sort&compress,times]{elsarticle}

\usepackage{enumerate}
\usepackage{amsthm}
\usepackage{amsmath}
\usepackage{amssymb}
\usepackage{xfrac}

\usepackage{tabularx,booktabs}
\newcolumntype{C}{>{\centering\arraybackslash}X} 
\newcolumntype{R}{>{$}r<{:$}} 

\newdefinition{rmk}{Remark}
\newproof{pf}{Proof}
\newproof{pot}{Proof of Theorem \ref{thm2}}

\newcommand*\diff{\mathop{}\!\mathrm{d}}
\newcommand*\MyExp{{\mathbb E}}

\usepackage{amsmath,amssymb,amsfonts}
\usepackage{algorithmic}
\usepackage{graphicx}
\usepackage{textcomp}
\usepackage{xcolor}
\def\BibTeX{{\rm B\kern-.05em{\sc i\kern-.025em b}\kern-.08em
		T\kern-.1667em\lower.7ex\hbox{E}\kern-.125emX}}

\usepackage[utf8]{inputenc}
\usepackage{times}
\usepackage{soul}
\usepackage{url}
\usepackage[hidelinks]{hyperref}
\usepackage[utf8]{inputenc}
\usepackage[small]{caption}
\usepackage{booktabs}
\usepackage{algorithm}

\usepackage{amsthm}

\newtheorem{theorem}{Theorem}

\usepackage{nomencl}
\makenomenclature

\newcount\Comments  
\usepackage{color}
\definecolor{darkgreen}{rgb}{0,0.5,0}
\definecolor{purple}{rgb}{1,0,1}
\newcommand{\kibitz}[2]{\ifnum\Comments=1\textcolor{#1}{#2}\fi}

\usepackage{hyperref}
\usepackage[absolute,overlay]{textpos}
\journal{Information and Software Technology}

\begin{document}
	\begin{textblock*}{20cm}(1cm,1cm)
		\textcolor{red}{Preprint accepted by the journal of Information and Software Technology (IST).
		\\
		DOI:	\url{https://doi.org/10.1016/j.infsof.2020.106393}}
	\end{textblock*}

	\begin{frontmatter}
		\title{Assessing Safety-Critical Systems from Operational Testing: A Study on Autonomous Vehicles}
		
		
		\author[1]{Xingyu Zhao}
		\ead{xingyu.zhao@hw.ac.uk}
		\author[2]{Kizito Salako}
		\ead{k.o.salako@city.ac.uk}
		\author[2]{Lorenzo Strigini}
		\ead{l.strigini@city.ac.uk}
		\author[1]{Valentin Robu}
		\ead{v.robu@hw.ac.uk}
		\author[1]{David Flynn}
		\ead{d.flynn@hw.ac.uk}
		\address[1]{The Smart System Group, School of Engineering and Physical Sciences,\\Heriot-Watt University, Edinburgh, EH14 4AS, United Kingdom}
		\address[2]{The Centre for Software Reliability, School of Mathematics, Computer Science and Engineering, \\City, University of London, Northampton Square, EC1V 0HB, United Kingdom}
		
		\begin{abstract}
		\textbf{Context}: Demonstrating high reliability and safety for  safety-critical systems (SCSs) remains a hard problem. Diverse evidence needs to be combined in a rigorous way: in particular, results of operational testing with other evidence from design and verification. Growing use of machine learning in SCSs, by precluding most established methods for gaining assurance, makes evidence from operational testing even more important for supporting safety and reliability claims.
		
		\noindent\textbf{Objective}: We revisit the problem of using operational testing to demonstrate high reliability. We use Autonomous Vehicles (AVs) as a current example. AVs are making their debut on public roads: methods for assessing whether an AV is safe enough are urgently needed. We demonstrate how to answer 5 questions that would arise in assessing an AV type, starting with those proposed by a highly-cited study.
		
		\noindent\textbf{Method}: We apply new theorems extending our Conservative Bayesian Inference (CBI) approach, which exploit the rigour of Bayesian methods while reducing the risk of involuntary misuse associated (we argue) with now-common applications of Bayesian inference; we define additional conditions needed for applying these methods to AVs.
		
		\noindent\textbf{Results}: Prior knowledge can bring substantial advantages if the AV design allows strong expectations of safety before road testing. 
		We also show how naive attempts at conservative assessment may lead to over-optimism instead; why extrapolating the trend of disengagements (take-overs by human drivers) is not suitable for safety claims; use of knowledge that an AV has moved to a ``less stressful'' environment.
		
		\noindent\textbf{Conclusion}: While some reliability targets will remain too high to be practically verifiable, our CBI approach removes a major source of doubt: it allows use of prior knowledge without inducing dangerously optimistic biases. For certain ranges of required reliability and prior beliefs, CBI thus supports feasible, sound arguments. Useful conservative claims can be derived from limited prior knowledge.
		\end{abstract}
		
		
	\end{frontmatter}

	\section{Introduction}
	
	Safety-Critical Systems (SCSs) play an important role in modern societies, with increasing numbers of applications in many domains like transportation, nuclear energy and  healthcare. How to assess SCSs that require very high reliability remains a challenging task after almost 30 years since the first publications \cite{littlewood_validation_1993,butler_infeasibility_1993} to highlight the problem. The main conclusions of \cite{littlewood_validation_1993} include: the reliability required from some systems is so high that gaining sufficient confidence in it from their failure-free operational test alone would require infeasible amounts of testing. These authors dubbed such requirements ``ultra-high reliability''. Combining operational testing evidence with evidence from other forms of verification may raise the level of reliability that can be validated. But \emph{how} to combine them, in practice, in a statistically principled way, to support high reliability claims remains an open question.
	
	In this paper, using Autonomous Vehicles (AVs) as our example, we revisit the problem of assessing SCSs with high reliability requirements. We also highlight some new challenges introduced by the use of Machine Learning (ML) in SCSs, and propose a new statistical inference method to address some major difficulties.
	

	Safety for conventional SCSs is guided by well-established industry standards, prescribed development processes, and verification techniques/tools that aid engineers build evidence as to whether a system is safe enough. However, the use of ML in safety critical applications calls for these to be revised \cite{bloomfield_disruptive_2019,alves_considerations_2018,burton_mind_2020}, to better reflect how ML approaches can make it even harder (compared to non-ML based systems) to estimate the probabilities of failures or accidents. An increased reliance on empirical demonstrations of safety and reliability via simulated (to the extent that such simulations can be trusted) and operational testing seems inevitable.
	
	Indeed, AV manufacturers have been testing their AVs on public roads in the U.S. for years: e.g., more than 20 million autonomous miles have been driven (and more than 10 billion autonomous miles simulated) by Waymo at the time of writing. Moreover, the amount of miles driven per year is increasing. Such operational testing in real traffic, with close observation of AV performance, has been important testimonial evidence in the U.S. Congress hearings on AV regulation \cite{urmson_hands_2016}. Meanwhile, various authors \cite{banerjee_hands_2018,kalra_driving_2016} have used the same kind of statistical data to draw sobering conclusions about how far AVs are from achieving their safety goals and (an even harder challenge \cite{littlewood_validation_1993,butler_infeasibility_1993}) demonstrating that these goals have been achieved before a vehicle type is accepted for routine autonomous operation.
	
	These studies mostly rely on descriptive statistics, giving insights on various aspects of AV safety \cite{banerjee_hands_2018,favaro_autonomous_2018,dixit_autonomous_2016,lv_analysis_2018}. A RAND Corporation study \cite{kalra_driving_2016} has been widely cited, and in this paper we refer to it for comparison, to illustrate similarities and differences between alternative statistical approaches to assessment and the results thereof. For the reader's convenience, we will refer to that paper as ``the RAND study''. The RAND study uses classical statistical inference to find how many miles need to be driven to claim a desired AV reliability with a certain confidence level. However, such techniques do not address how safety and reliability claims\footnote{In this paper we only deal with probabilistic claims, so ``reliability'' claims will be about probabilities of occurrence of failures, ``safety'' claims about failures that are safety-relevant. The two kinds do not require different statistical reasoning, except as far as affected by practical differences in e.g., frequencies, desired bounds, and degrees of observability.}, based on operational testing evidence, can be made in a way that:
	
	\textit{a) is practical given very rare failure events}, such as fatalities and crashes. If and when AVs achieve their likely safety targets, rates of such events will be very small, say a $10^{-10}$  \textit{probability of a fatality event per mile} (\textit{pfm}). 
	Gaining confidence in such low failure rates is challenging \cite{littlewood_validation_1993,butler_infeasibility_1993}, possibly requiring infeasible amounts of \textit{failure-free} operation to discriminate between the conjectures that the \textit{pfm}, for instance, is as low as desired or not.
	If some accidents do occur, as is the case, the original challenges found in \cite{littlewood_validation_1993,butler_infeasibility_1993} become even harder. This was the case in the RAND study findings.
	
	\textit{b) incorporates relevant prior knowledge}. In conventional systems, such prior knowledge would typically include evidence of soundness of design (as supported by verification results) and quality of process. AVs rely on ML software for core functionality, and the ability to prove correct design is lacking (despite intense research). But AVs, just as more conventional systems, will normally include safety precautions: e.g., defence-in-depth design with safety monitors/watchdogs. 
	Indeed, such ``safety subsystems'' are not only recommended in policy documents \cite{anderson_autonomous_2016,safety_first}, but also extensively implemented by AV manufacturers \cite{waymo_waymo_2018,amnon_shashua_plan_2017}. These subsystems have relatively simple functionality (e.g., bringing the vehicle to a safe stop) and could possibly avoid relying on ML functions, thus allowing conventional verification methods. If such subsystems form the basis for prior confidence in safety, evidence about their development and verification should be combined (in a statistically principled way) with operational testing evidence. The same applies if safety evidence for the ML functions (e.g., from automated testing and formal verification of Neural Networks \cite{tian_deeptest_2018,huang_safety_2017}) and for the whole system (e.g., with a more direct matching between architecture, verification methods and arguments) \cite{fisher_verifying_2013,fisher_verifiable_2018,	bloomfield_disruptive_2019,koopman_credible_2019}) is available.
	
	
	\textit{c) considers that while road testing data is collected, the AVs undergo updating and are deployed in different environments}. For an unchanging vehicle that operates under statistically unchanging conditions, ``constant event rate'' models, as applied, e.g., in the RAND study, may apply. However, there is an expectation that AV safety improves as the AV evolves (i.e. its ML-based core systems ``learn'') with driving experience, or that the AV is deployed in different environments with different road/traffic conditions, and both kinds of change will affect the frequency of failures.
	
	The present paper is an extension of our conference paper \cite{zhao_assessing_2019}, with new content listed as the last 3 contributions in the following list. The key contributions of this work are: 
	
	\textit{1)} For constant failure rate scenarios, we develop a new \emph{Conservative Bayesian Inference} (CBI) method for reliability assessment, that can incorporate both failure-free and ``rare failures'' evidence. For AVs, occasional failures \emph{are} to be expected. Including operational evidence with ``rare failures'' into the assessment generalises existing CBI methods (applied in other settings such as nuclear safety) that, so far, consider only failure-free evidence \cite{bishop_toward_2011,strigini_software_2013,zhao_modeling_2017,zhao_conservative_2015,zhao_conservative_2018}. 
	Being a Bayesian approach, CBI allows for the incorporation of prior knowledge of non-road-testing evidence (e.g., verified aspects of the behaviour of an AV’s ML algorithms; verification results for the safety subsystems). We then compare claims based on our CBI framework with claims from other AV studies, using the same data and settings (in particular, we consider how CBI compares with the well-known inference approach used in the RAND study). CBI shows how these other approaches can be either optimistic, or too pessimistic, and the difference may be substantial.
	
	\textit{2)} For assessing changes in failure rate, we extend CBI to statistical inference using \textit{bivariate} prior distributions, so that partial prior knowledge on the relationship between the (unknown) failure rates before and after the changes can be used to answer practical questions regarding the deployment of a new version of an AV, or deployment of the same AV in a new environment. To the best of our knowledge, this is the first work to formalise such questions about AVs using a statistical model.
	 
	\textit{3)} In practice, assessors may be interested in different reliability measures, e.g., expected failure rate or confidence bounds on a required failure rate. The meaning of ``being conservative'' varies as the reliability measure under study changes. By way of numerical examples,  we exemplify errors that can occur if the relationship between reliability measures and conservatism is misunderstood.
	
	\textit{4)} In the original conference paper \cite{zhao_assessing_2019}, we showed how past AV disengagement\footnote{Events in which AVs' control is switched to human drivers, e.g. due to failures.} data can be used by \emph{Software Reliability Growth Models (SRGMs)} \cite{Miller1986EOS} to predict future \textit{disengagement per mile} (\textit{dpm}), and emphasised that using SRGMs to predict \textit{dpm} is a valuable tool for planning (but not safety assessment). The present paper provides more complete arguments: against basing safety decisions directly on statistical extrapolations (e.g., via SRGMs) of the trend of \textit{dpm}; and about how SRGM-based arguments could be made  relevant.

	
	The outline of the rest of this paper is as follows. In section \ref{sec_OT_and_failure_process}, we present preliminaries on assessing reliability from operational testing. Section \ref{sec_CBI} and \ref{sec_CBI_2d} then detail the new CBI methods for scenarios of constant, and then of changing, failure rate. Sections \ref{sec_fallacy_be_conservative} and \ref{sec_waring_dpm_srgm} discuss (and illustrate) risks with uninformed attempts at conservative assessments, and with using extrapolations of disengagement trends \textit{dpm} prediction for safety claims. Finally, sections \ref{sec_related_work} and \ref{sec_conclusions} summarise related work, contributions and future work.
	
	\section{Operational testing \& failure processes}
	\label{sec_OT_and_failure_process}
	For conventional SCSs, statistical evaluation from operational testing, or ``proven in use'' arguments, are part of standards like IEC61508 \cite{iec_61508_2010} and EN50129 \cite{cenelec_bs_2018}. These practices are supported by probabilistic methods, both established \cite{atwood2003handbook,strigini_guidelines_1997,may_reliability_1995} and still evolving  \cite{walter_bayesian_2017,bishop_deriving_2017,utkin_imprecise_2018}. Since, for AVs, road testing is emphasised as evidence for building public confidence in safety and reliability,
	inference methods using such operational evidence should indeed attract attention.

	In general, depending on the system under study, a stochastic failure process is chosen as a mathematical abstraction of reality. Here, for AVs, we describe the failure processes -- for the occurrence of fatalities or crashes -- as \textit{Bernoulli processes}. These models assume the probability of a failure\footnote{For brevity, we call ``failure'' generically the event of interest (fatality, crash, etc.), and use ``failure rate'' both in its technical meaning as the parameter (\textit{dpm}) of, say, a Poisson process, and for the probability of failure per mile in a Bernoulli model (e.g., for \textit{pfm}).} per driven mile is a constant, and events in one mile are independent of events in any other mile driven. This process assumption may not really hold for various reasons that depend on the contexts. But in many practical scenarios a Bernoulli model is an acceptable approximation of the more complex, real process.

	\textit{a)} For constant failure rate scenarios, we assume a ``finalised'' version of the AV deployed in 
	unchanging environmental conditions. 
	In practice, AVs are typically updated while road testing progresses.
	A possible argument for still using a constant failure rate model as a first approximation could be, for instance, that the non-ML based safety subsystems makes the failure rate for the overall AV much smaller than that of the ML-based systems alone, and this overall AV reliability remains constant during observation, despite the online evolution of the ML-based systems, or the small changes of road conditions from a safety perspective\footnote{``A first approximation'' because the evolution of the ML-based core changes the set of failures to be tolerated by the safety subsystem (cf. \cite{PopovStrigini2010ISSRE}). A previous statistical study \cite{favaro_examining_2017} found that some key AV reliability measures, e.g. the accident rate for AVs, appear constant over time. But this is not enough to support making it a modelling \emph{assumption}.}. Thus, there are two reasons for us to use this model: i) the model is simple enough to highlight the challenges of AV safety assessment, and ii) for the purpose of comparison against the RAND study \cite{kalra_driving_2016} which uses this model.
	


	\textit{b)} On the other hand, we also consider scenarios with a change of failure rate, in which we assume there is a significant version update of the AV (e.g., a new software architecture or a thoroughly retrained ML component) or a non-negligible change of environmental conditions (e.g., moving to operation in another country). Then, the probabilities of the failures that a safety subsystem will mitigate, and those that it cannot mitigate, will change. This, in turn, could have a notable effect on the safety of the AV. In this case, we still assume the failure processes of the AV before and after the changes as Bernoulli models, for the reasons discussed above, while the statistical inference is done on a bi-variate probability distribution of the unknown failure rates.


	\section{The CBI as a constant event-rate model}
	\label{sec_CBI}

	Assessment claims using statistical inference come in different flavours. The RAND study derives ``classical'' confidence statements about the claim of an acceptable failure rate. For instance, $95\%$ confidence in a bound of $10^{-x}$ means that if the failure rate were greater than $10^{-x}$, the chances of observing no failures in the miles driven would be $5\%$ at most. This quantifies the extent to which the empirical test (of that many miles of road testing) challenges an unreliable system, and is often used for deciding whether to accept the system for operation. The Bayesian approach, instead, treats failure rate as a random variable with a ``prior'' probability distribution (``prior'' to test observations). The prior is updated (via Bayes' theorem) using test results, giving a ``posterior'' distribution. Decisions are based on probabilities derived from the posterior distribution, e.g., the probability (``Bayesian confidence''), say 0.95, of the failure rate being less than $10^{-x}$. These two notions of confidence have radically different meanings, but
	decision making based on levels of ``confidence'' of either kind is common: hence we will compare the amounts and kinds of evidence required to achieve high ``confidence'' with either approach.
	
	Now, a challenge for using Bayesian inference in practice is the need for complete prior distributions (of the failure rate, in the present problem).
	A common way to deal with this issue is to choose distribution functions that seem plausible in the domain and/or mathematically convenient (e.g. for conjugacy). 
	However, forcing oneself to state such a complete distribution may well mean that the distribution itself does not describe only one's prior knowledge, but adds extra, unjustified assumptions. This may do no harm if the posterior depends on the data much more than on the prior distribution, but in our case (with possibly zero failures), 
	the conclusions of the inference will be seriously sensitive to these assumptions: those extra assumptions risk dangerously unsound reasoning. 
	
	
	CBI bypasses this problem: rather than a \textit{complete} prior distribution, an assessor is more likely to have (and be able to justify) more limited \textit{partial prior knowledge}, e.g. a prior confidence bound -- ``I am 80\% confident that the failure rate is smaller than $10^{-3}$'' -- based on e.g. experience with results of similar quality practices in similar projects.
	This partial prior knowledge is far from a complete prior distribution. Rather, it \textit{constrains} the prior: there is an \textit{infinite set} of prior distributions satisfying the constraints. Then, depending on the specific reliability measure of interest (e.g., posterior expected failure rate or a posterior confidence bound on a required failure rate -- the example we focus on in this paper), CBI seeks a prior distribution, within the set of \emph{all} prior distributions satisfying the partial prior knowledge, that gives the \textit{most conservative} result for the posterior prediction.

    The essential idea of CBI is applicable in a variety of contexts and scenarios \cite{bishop_toward_2011,strigini_software_2013,zhao_conservative_2015,zhao_modeling_2017,zhao_conservative_2018}. It has been investigated for various objective functions (the posterior measures of interest) with different forms of constraints (the partial prior knowledge), e.g., a posterior expected failure rate given a prior confidence bound in \cite{bishop_toward_2011}. However, all published CBI methods are for conventional SCSs (e.g., nuclear protection systems, where any failure is assumed to have significant consequences), and thus deal with operational testing where \textit{no} failures occur. AI systems do fail in operation. For AVs, crashes and fatalities, although rare, have been reported. To deal with (infrequent) failures, we propose a more general CBI method, in which $0$ failures becomes a special case. This is reflected by a more general form of the likelihood function in the Bayesian inference.

    For AVs, here we apply CBI to assessing \textit{pfm} and \textit{probability of seeing crash-events per mile} (\textit{pcm}). To compare the results with those of the RAND study, the new CBI theorems we present use as objective function the probability of the \textit{pfm} (or \textit{pcm}) being smaller than a required bound after seeing road testing evidence.


	\subsection{CBI with failures in testing}
	
	As described in Section \ref{sec_OT_and_failure_process}, consider a Bernoulli process representing a succession of miles driven by an AV, and let $X$ be the unknown \emph{pfm} value (the setup if one considers crashes instead of fatalities is analogous). Suppose $k$ failures in $n$ driven miles are observed (denoted as $k\&n$ in the equations, for brevity). If $F(x)$ is a prior distribution function for $X$ then, for some stated reliability bound $p$,
	\begin{equation}
	\label{eq_post_cf_bound_with_complete_prior}
	Pr(X \leqslant p\mid k\&n)=\frac{\int_{0}^{p} x^k(1-x)^{n-k} \mathrm dF(x) }{\int_{0}^{1} x^k(1-x)^{n-k} \mathrm dF(x)}
	\end{equation}

	As an example, suppose that, rather than some complete prior distribution, only partial prior beliefs are expressed about an AV's \emph{pfm}:
	\begin{equation}
	\label{eq_prior_constraints_1}
	Pr(X \leqslant \epsilon)=\theta,\quad Pr(X\geqslant p_l)=1
	\end{equation} 
	The interpretations of the model parameters are:
	
	$\bullet \enspace \epsilon$ is the engineering goal: a target level that developers try to achieve for a reliability or safety measure (e.g. \textit{pfm}). To illustrate, for \textit{pfm}, this goal could be two \cite{liu_how_2019}, or three \cite{amnon_shashua_plan_2017}, orders of magnitude safer than the average for human drivers.
	
	
	$\bullet \enspace \theta$ is the prior confidence (\emph{before} testing the AV on public roads) that the engineering goal has been achieved. Such prior confidence would have to be high enough to decide to proceed to testing on public roads. It could be obtained from simulations, verification of the AV safety subsystems, etc. For instance, high quality development and verification of correctness against rigorously verified requirements would give some confidence that these subsystems are fault-free, or ``perfect'' or approximately so (as discussed more in depth elsewhere, e.g., \cite{BertolinoStriginiSTVR1998perfection,strigini_software_2013,zhao_modeling_2017}).
	An initial value $\theta$ for this confidence would derive from historical evidence 
	(essentially: what fraction of similar systems, similarly proved to be free of safety faults, were actually so, as far as is known after extensive operation?), tempered with some prudence about the validity of the data.
	Failure free simulated operation would strengthen this confidence \cite{zhao_modeling_2017}. 
	
	$\bullet \enspace p_l$ is a lower bound on the failure rate: the best reliability claim feasible given current vehicle technology. 
	For instance, \textit{pfm} cannot be smaller than, say $10^{-15}$, due e.g. to the possibility of catastrophic hardware failures  (tyre/engine fails on a highway), \emph{even if the AV's control and safety systems, including the ML parts, were perfect}.
	$p_l$ would be estimated from historical statistics of such accidents. 
	
	
	We note that $\epsilon>p_l$ because even extensive historical evidence of efficacy of verification could not discriminate between systems that are indeed free of design faults and systems where design faults only exist that cause very low failure rates \cite{strigini_software_2013,zhao_conservative_2015}. 
	
	
We have outlined how the values of parameters in (\ref{eq_prior_constraints_1}) could be chosen in practice to demonstrate that (\ref{eq_prior_constraints_1}) is a plausible form for prior beliefs that can be supported with reasonable arguments. Other forms may be found, with different parameters, depending on the evidence available; or other interpretations could be applied to the parameters in (\ref{eq_prior_constraints_1}), varying between manufacturers and across business models. 

\begin{figure}[htbp!]
		\centering
		\includegraphics[width=1\linewidth]{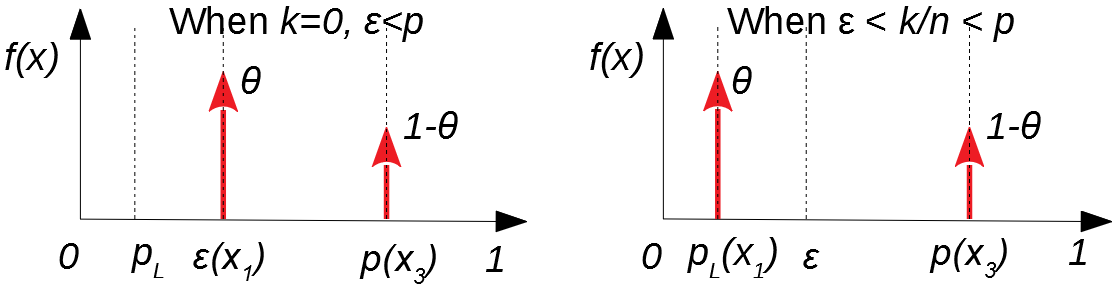}
		\caption{Conservative two-point priors for two choices of model parameters -- with failure free data (left) and rare failures (right).}
		\label{fig_two_point_priors}
	\end{figure}

	Now, assuming one has the prior beliefs \eqref{eq_prior_constraints_1}, the following CBI theorem shows what these beliefs allow one to rigorously claim about an AV's safety and reliability.
	\begin{theorem} A prior distribution that gives the infimum for \eqref{eq_post_cf_bound_with_complete_prior}, subject to the constraints \eqref{eq_prior_constraints_1}, is the two-point distribution $Pr(X=x)=\theta{\bf 1}_{x=x_1} + (1-\theta){\bf 1}_{x=x_3}\,$, where $p_l\leqslant x_1 \leqslant \epsilon < x_3\,$ and the values of $x_1$, $x_3$ both depend on the model parameters (i.e. $p_l, \epsilon, p$) as well as $k$ and $n$. Using this prior, the infimum for \eqref{eq_post_cf_bound_with_complete_prior} is
		\allowdisplaybreaks\begin{align}
		\frac{x_1^k(1-x_1)^{n-k}\theta}{x_1^k(1-x_1)^{n-k}\theta + x_3^k(1-x_3)^{n-k}(1-\theta)}{\bf 1}_{p>\epsilon}
		\label{eq_CBI_post_conf_bound_see_failures}
		\end{align}
		where ${\bf 1}_{\tt S}$ is an indicator function -- it is equal to 1 when {\tt S} is true and 0 otherwise.
		\label{thrm_1}
	\end{theorem}
	The proof of Theorem 1 is in \ref{sec_app_A}. Depicted in Fig.~\ref{fig_two_point_priors} are two common situations (given different values of the model parameters): with failure-free and ``rare failures'' evidence.
	
	Solving \eqref{eq_CBI_post_conf_bound_see_failures} for $n$ -- the miles to be driven to claim the \emph{pfm} is less than $p$ with probability $c$, after seeing $k$ failures -- provides our main technical result. From a Bayesian perspective, $n$ will depend on the prior knowledge \eqref{eq_prior_constraints_1}. In what follows, we compare the proposed $n$ values from CBI, the RAND study, a Uniform prior and Jeffreys prior (as suggested by regulatory guidance like \cite{atwood2003handbook}). Similar comparisons can be made for \textit{pcm}; we omit these due to page limitations.
	
	\subsection{Numerical examples of CBI for \textit{pfm} claims}
	\label{sec_num_examples_CBI}
	
	In the RAND study, data from the U.S. department of transportation supported a \textit{pfm} for human drivers of $1.09e{-8}$ in 2013. For illustration, suppose that a company aims to build AVs two orders of magnitude safer, i.e. $\epsilon=1.09e{-10}$, as proposed by \cite{liu_how_2019}. Also, assume $p_l=10^{-15}$: that is, the unknown \textit{pfm} value cannot be better than $10^{-15}$.
	
	\textbf{Q1: How many fatality-free miles need to be driven to claim a \textit{pfm} bound at some confidence level?}
	
\begin{figure}[htbp!]
		\centering
		\includegraphics[width=1\linewidth]{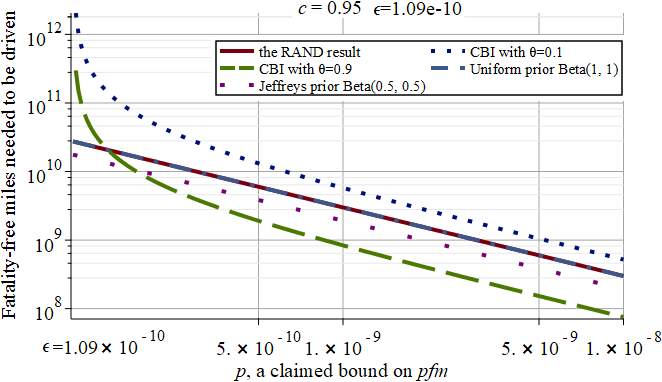}
		\caption{Fatality-free miles needed to be driven to demonstrate a \textit{pfm} claim with $95\%$ confidence. \emph{Note}: the curves for Bayes with a uniform prior and the RAND results overlap; to be precise, there is a constant difference of 1 between them, which is simply a consequence of the similarity between their analytical expressions in this scenario).}
		\label{fig_pfm_no_failure}
	\end{figure}

	With the prior knowledge \eqref{eq_prior_constraints_1}, we answer Q1 by setting $k=0$ and solving \eqref{eq_CBI_post_conf_bound_see_failures} for $n$. Fig.~\ref{fig_pfm_no_failure} shows the CBI results with $\theta\!=\!0.1$ (weak belief) and $\theta\!=\!0.9$ (strong belief) respectively, compared with the RAND results, and Bayesian results with a uniform prior {\tt Beta}$(1,1)$ and the Jeffreys prior for Binomial models ({\tt Beta}$(0.5,0.5)$ \cite[p.6.37]{atwood2003handbook}).
	Fig.~\ref{fig_pfm_no_failure} shows that \eqref{eq_CBI_post_conf_bound_see_failures} can imply significantly more, or less, miles must be driven than suggested by either the RAND study or the other Bayesian priors -- depending on how confident one is \emph{before seeing test results} that the goal $\epsilon$ has been reached. For instance, to claim, with 95\% confidence, that AVs are as safe as human drivers (so $p=1.09e{-8}$), the RAND analysis requires 275 million fatality-free miles, whilst CBI with $\theta=0.9$ only requires 69 million fatality-free miles, with 90\% prior confidence that the AVs are two orders of magnitude safer than humans (based on, e.g., having the core ML-based systems backed up by non-ML safety channels that are relatively simple and easier to be verified. Such verification can be the case in traditional SCSs \cite{littlewood_reasoning_2012}).
	
	Alternatively, if one has only a ``weak'' prior belief in the engineering goal being met ($\theta=0.1$), then CBI requires 476 million fatality-free miles -- significantly more than the other approaches compared.
	
	The reader should not be surprised that our conservative approach does not always prescribe more fatality-free miles be driven than that prescribed by the RAND study -- different decision criteria and statistical inference methods can yield different results from the same data \cite{berger_could_2003}. However, it is true that, for any confidence $c$, CBI will require significantly more miles than the RAND study prescriptions for all claims $p$ ``close enough'' to the engineering goal $\epsilon$.
	
\begin{figure}[htbp!]
		\centering
		\includegraphics[width=1\linewidth]{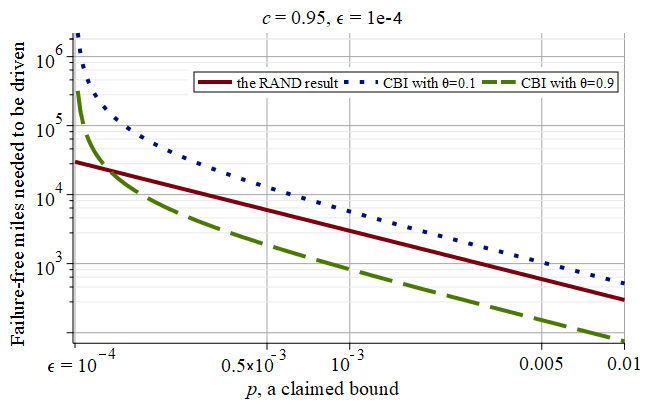}
		\caption{Failure-free miles needed to be driven to demonstrate a less stringent reliability claim with $95\%$ confidence.}
		\label{fig_less_stringent_claim}
	\end{figure}

	We note that, for AVs that may have less stringent reliability requirements (e.g. AVs for industrial/agricultural use in restricted environments),
	both the engineering goal and reliability claims can be much less stringent (i.e., higher) than the examples in Fig.~\ref{fig_pfm_no_failure}. 
	For such a scenario,
	Fig.~\ref{fig_less_stringent_claim} shows our CBI results alongside those from the RAND study's approach, given an engineering goal $\epsilon=10^{-4}$ and a range $[10^{-4} , 10^{-2}]$ for the claimed bound $p$. Although it shows the same pattern as Fig.~\ref{fig_pfm_no_failure}, the evidence required to demonstrate a reliability claim being met with the given confidence level is much less and within a feasible range. For instance, when the claim of interest is $p=10^{-3}$, CBI with a strong prior belief in the engineering goal being met (i.e. $\theta=0.9$) requires less than $10^3$ failure-free miles, while the RAND method requires 2 to 3 times as many.
	
	Notice that, for all of the scenarios we have presented so far, no amount of testing will support trust in any bound $p$ lower than $\epsilon$. This is because of constraint \eqref{eq_prior_constraints_1}. It allows a range of possible prior distributions -- and thus posterior confidence bounds -- but, as our theorems show, it gives no basis for trusting any bound better than $\epsilon$ (as exemplified in Fig.~\ref{fig_pfm_no_failure}). Hence, a conservative decision maker that has partial prior knowledge \eqref{eq_prior_constraints_1} 
	cannot accept a claim, on the basis of the fatality-free operation, that the AV reliability exceeds the engineering goal.
	
	This consistency with the limited beliefs that one can confidently bring to the inference is the strength of CBI. We based the CBI example in this paper on the form of beliefs (\ref{eq_prior_constraints_1}), which we think can be reasonably argued in practice (not necessarily for the $\epsilon$-level in Fig.~\ref{fig_pfm_no_failure}); but
if \emph{further} evidence justified a prior belief in some bound $p$ ($<\epsilon$), this further constraint on the set of possible priors would cause CBI to give less pessimistic claims.

	\textbf{Q2: How many miles need to be driven, with fatality events, to claim a \emph{pfm} bound at some confidence level?}

	The RAND study answers this question via classical hypothesis testing, choosing as an example a confidence bound 20\% better than human drivers' \textit{pfm} in 2013.
	Their result (in number of miles required) is shown in boldface in Table.~\ref{tab_miles_to_drive_with_failures}.
	
	In the Bayesian approach, posterior confidence depends on observations. In order to compare with the RAND study result, we thus postulate an observed number of fatalities consistent with the RAND study analysis. As an example, we consider that, given a \textit{pfm} equal to the above confidence bound, and driving the number of miles found necessary in the RAND study, the expected number of fatalities would be
	$k=8.72e{-9} \times 4.97e9 \approx 43$ (where $8.72e{-9}$ is a reliability claim obtained from $4.97e9$ fatality free miles in the RAND model).
	We thus assume 43 fatalities and show in column 1 of Table~\ref{tab_miles_to_drive_with_failures} the miles required by the Bayesian approaches, including CBI, Uniform and Jeffreys priors. In addition to the purpose of comparison, this case also represents a long term scenario in which, as popularity and public use of AVs grow, the count of fatal accidents progressively reaches high values. 
	We show what evidence would then be needed to reassure the public that reliability claims are still being met.
	
	For a short term scenario, as a second example, the last column of Table~\ref{tab_miles_to_drive_with_failures} shows the corresponding results, if only one fatality occurs. Again, we compare the results of classical hypothesis testing, CBI and using other Bayesian priors.
	
	
	
	All of the examples in Table~\ref{tab_miles_to_drive_with_failures} ``agree'': the miles needed to make these claims are prohibitively high. However, given the  prior beliefs we assume for CBI, the CBI numbers \emph{require 10$\sim$20 times more miles than the rest if 43 fatalities are seen}. The number at the bottom of column 1 represents the miles needed to demonstrate that, after fatalities consistent with \emph{pfm}$=8.72 e{-9}$, there is only a $5\%$ chance of the true \textit{pfm} being worse than that.
	The difference from the RAND results may seem large, but it is in the interest of public safety: CBI avoids any implicit, unwittingly optimistic assumptions in the prior distribution.
	We recall that with no fatalities, the CBI example \textit{does} offer a sound basis for achieving high confidence with substantially fewer test miles than the RAND approach requires (e.g. 69 \textit{vs} 275 million miles).

	\begin{table}[htbp!]
		\centering
		\resizebox{0.9\columnwidth}{!}{%
			\begin{tabular}{l|c|c}  
				\toprule
				&\textit{p}=8.72e-9, \textit{k}=43 & \textit{p}=4.12e-9, \textit{k}=1\\
				\midrule
				Classical & $\mbox{\bf 4.97e9}$  & $2.43e8$      \\
				Uniform priors & $6.40e9$  &   $1.15e{9}$\\
				Jeffreys priors & $6.33e9$  &  $9.48e8$ \\
				CBI with $\theta=0.9$  & $7.89e10$  & $3.88e9 $     \\
				\bottomrule
			\end{tabular}
		}
		\caption{Miles needed to support a \emph{pfm} claim $p$ with 95\% confidence, with $k$ fatalities.}
		\label{tab_miles_to_drive_with_failures}
	\end{table}
	
	\textbf{Q3: How many more fatality-free miles need to be driven to compensate for one newly observed fatality?}
	
	This question relates to the following plausible scenario.
	An AV has been driven for $n_1$ fatality-free miles, justifying a \textit{pfm} claim, say $p$ (with a fixed confidence $c$), via CBI based on this evidence and some given prior knowledge. Then suddenly a fatality event happens. Instead of redesigning the system (as no evidence exists to point to a technical/AI control design fault), the company still believes in its prior knowledge, attributes the fatality to ``bad luck'', and asks to be allowed more testing to prove its point.
	If  the public/regulators accept this request,
	it is useful to know how many extra fatality-free miles, say $n_2$, are needed to compensate for the fatality event, so that the company can demonstrate the same reliability $p$ with confidence $c$.
	
\begin{figure}[bhtp!]
		\centering
		\includegraphics[width=1\linewidth]{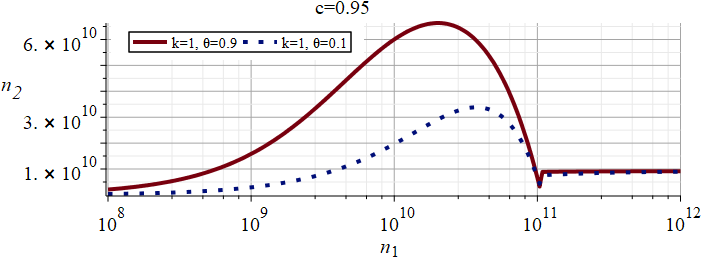}
		\caption{Fatality-free miles needed to compensate one newly observed fatality given $n_1$ fatality-free miles has been driven before.}
		\label{fig_extra_miles_needed}
	\end{figure}

	To answer this, apply the CBI model in two steps (fixing the confidence level $c$ and prior knowledge $\theta$): (i) determine the claim $[ X\!\leqslant\!p\,]$ that $n_1$ will support with $k\!=\!0$ (i.e. fix $k,n$ \& solve \eqref{eq_CBI_post_conf_bound_see_failures} for $p$). (ii) determine the miles that support the claim $[X\!\leqslant\!p\,]$ upon seeing $k\!=\!1$ (i.e. fix $k,p$ \& solve \eqref{eq_CBI_post_conf_bound_see_failures} for $n$). Then $n_2 \!=\! n - n_1$ more fatality-free miles are needed to compensate for the fatality; we plot some scenarios in Fig.~\ref{fig_extra_miles_needed}.

	The solid curve in Fig.~\ref{fig_extra_miles_needed} shows a uni-modal pattern, decreasing as $n_1$ approaches the value $n^\ast=1.06e{11}$ (with a corresponding $p$ value, $p^\ast=1.16e{-10}$, derived from the 1st step), then increasing again with an asymptote of $n_2=1/\epsilon$, as $n_1$ goes to infinity. A complete formal analysis deriving $p^\ast$ and the asymptote of $n_2=1/\epsilon$ is in \ref{sec_app_B}.
	
	Intuitively, the more fatality-free miles were driven, the higher one's confidence in reliability; and thus, the more miles needed to restore that confidence after a fatality occurs. But, if $n_1$ was so high as to allow confidence in a claim close to $p^\ast$, then after the fatality, a much smaller $n_2$ is needed to be able to claim $p^\ast$ again. As $n_1$ tends to infinity, interestingly, there is a ceiling on the required $n_2$, \textit{for all values} of $c$ and $\theta$. We note that the shape of the curve (including the asymptote on the right) is invariant with respect to $c$ and $\theta$.

	%
	%
	%
	%
	\section{CBI for changing event rate}
	\label{sec_CBI_2d}
	
	In the previous section, we assumed an unchanging vehicle (in terms of \textit{pfm}) operating under environments that have unchanging statistical properties, and thus a ``constant event rate'' CBI model. In this section, we consider  scenarios in which the event rate of interest could change between the ``testing'' regime and the regime for which a prediction is sought. For instance, one might reasonably expect changes in event rates if future use of the AV is in different climates or regions from the testing (because these could imply different frequencies of weather conditions -- like thunder storms vs sunny spells -- or of road works, or heavy and light traffic conditions), or different seasons (e.g, testing  in summer for predicting rates in the following winter). The analyses presented in this section are a starting point for conservative assessment under such situations. By way of example, we will refer to the following two scenarios in the discussion that follows:
	\begin{itemize}
		\item \textbf{Q4}: the AV has been tested on the roads in City-A for $n_A$ fatality-free miles. Now the company wants to deploy the AV to City-B. We have high confidence (say $\phi$) that the road conditions of the two cities are similar and the change of environments should not harm safety. However, to be conservative, how many new fatality-free miles need to be driven in City-B (denoted as $n_B$) to claim a required \textit{pfm} bound for City-B, say $p_B$, with a given confidence level $c$?
		\item \textbf{Q5}: Version-A of the AV has been tested extensively, say for $n_A$ fatality-free miles, on public roads. Now we have updated the AV to a new Version-B. We have high confidence (say $\phi$) that the \textit{pfm} of Version-B should be no worse than that of Version-A. However, to be conservative, how many extra fatality-free miles need to be driven for the Version-B (denoted as $n_B$) to claim a required \textit{pfm} bound, say $p_B$, with a given confidence level $c$?
	\end{itemize}

	To answer the questions in the above scenarios, we develop a new CBI model with two variables, $X$ and $Y$, representing respectively the unknown \textit{pfm} values of the two cities/versions -- $\mathit{pfm}_A$ and $\mathit{pfm}_B$. Thus, instead of a one-dimensional prior distribution $F(x)$ as in Eq.~\eqref{eq_post_cf_bound_with_complete_prior}, there is now a two-dimensional joint prior distribution $F_{AB}(x,y)$. Then, for some required bound $p_B$, our objective function -- the posterior confidence in the bound after seeing $n_A$ and $n_B$ fatality-free miles of the two cities/versions is:
	\begin{equation}
	\label{eq_post_conf_2D}
	Pr(Y\leqslant p_B \mid n_A, n_B)=
	\frac{\int_0^{p_B}\!\int_0^1 (1-x)^{n_A}(1-y)^{n_B} \mathrm dF_{AB}(x,y) }{\int_0^{1}\!\int_0^1 (1-x)^{n_A}(1-y)^{n_B} \mathrm dF_{AB}(x,y)}
	\end{equation}
	
\begin{figure}[bhtp!]
		\centering
		\includegraphics[width=0.7\linewidth]{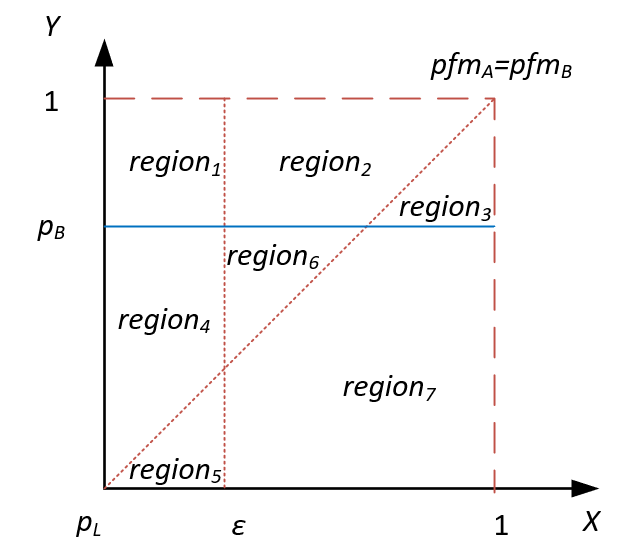}
		\caption{The sample space on which the joint prior distribution $F_{AB}(x,y)$ is defined, with 7 regions of interest. A distribution $F_{AB}(x,y)$ associates a probability mass $M_i$ to each region$_i$.}
		\label{fig_joint_prior_fab}
	\end{figure}

	The CBI philosophy is even more appropriate in this case: it is even harder for assessors to have a complete bivariate prior distribution; rather, they will have some limited partial knowledge about it. To deal with the Q4 and Q5 scenarios, we consider a joint prior distribution $F_{AB}(x,y)$ defined over the unit-square in Fig.~\ref{fig_joint_prior_fab}, where 7 regions of interest appear (each region$_i$ to be associated with a probability mass $M_i$) and:
	\begin{itemize}
		\item for City-A/Version-A, we have, as in our previous scenario, marginal partial knowledge of $\mathit{pfm}_A$ as shown in Eq.~\eqref{eq_prior_constraints_1}: a certain lower bound $p_l$ and $\sum_{i=1,4,5} M_i=\theta$.
		\item for the new City-B/Version-B, we have 
		\begin{equation}
		\label{eq_nwtes_certain_lowerB}
		Pr(Y\leqslant X)=\phi , \quad Pr(Y\geqslant p_l)=1
		\end{equation}
		That is, confidence $\phi$ that the $\mathit{pfm}_B$ is no worse than the $\mathit{pfm}_A$ (i.e. $\sum_{i=3,5,7} M_i=\phi$). Also, just as for City/Version-A, there is a lower bound $p_l$ on $\mathit{pfm}_B$.
	\end{itemize}

\begin{figure}[bhtp!]
		\centering
		\includegraphics[width=0.7\linewidth]{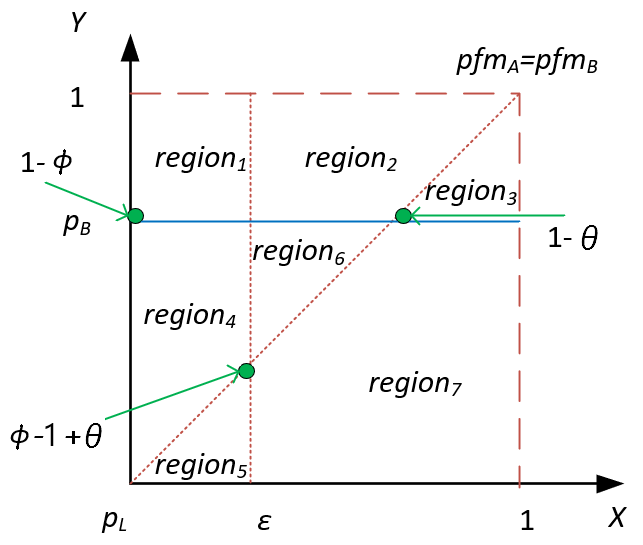}
		\caption{The worst-case 3-point joint prior distribution that gives the infimum for Eq.~\eqref{eq_post_conf_2D} when $\phi>1-\theta$, subject to constraints \eqref{eq_prior_constraints_1} and \eqref{eq_nwtes_certain_lowerB}.}
		\label{fig_worst_case_joint_prior}
	\end{figure}

	\begin{theorem}
		\label{theorem_two_d_CBI}
		A prior distribution that gives the infimum for \eqref{eq_post_conf_2D}, subject to constraints \eqref{eq_prior_constraints_1} and \eqref{eq_nwtes_certain_lowerB}, is a three-point distribution. 
		When $\phi>1-\theta$, as shown in Fig.~\ref{fig_worst_case_joint_prior}, the prior is $Pr(X=x,Y=y)=(1-\phi){\bf 1}_{x=p_l,y=p_B} + (1-\theta){\bf 1}_{x=p_B,y=p_B}+(\phi-1+\theta){\bf 1}_{x=\epsilon,y=\epsilon}$. Using this prior, the infimum for \eqref{eq_post_conf_2D} is
		\begin{equation}
		\label{eq_worst_case_2d_posterior}
		\frac{(1-\epsilon)^{n_A+n_B}M_5}
		{(1\!-\!\epsilon)^{n_A+n_B}M_5\!+\!(1\!-\!p_B)^{n_A+n_B}M_3\!+\!(1\!-\!p_l)^{n_A}(1\!-\!p_B)^{n_B}M_1} {\bf 1}_{\phi>1-\theta}	
		\end{equation}
		where $M_1=1-\phi$, $M_3=1-\theta$, $M_5=\phi-1+\theta$ and ${\bf 1}_{\tt S}$ is an indicator function -- it is equal to 1 when {\tt S} is true and 0 otherwise.
		When $\phi \leqslant 1-\theta$, the worst-case prior distribution will always yield 0 as the infimum for \eqref{eq_post_conf_2D}.
	\end{theorem}
	The proof of Theorem \ref{theorem_two_d_CBI} is in \ref{sec_app_C}. Given a required level of confidence, say $c=95\%$, and other model parameters (i.e. $p_B$, $p_l$, $\epsilon$, $\phi$ and $\theta$), we solve Eq.~\eqref{eq_worst_case_2d_posterior} for $n_B$, obtaining the answer we are seeking for Q4 and Q5 (the relevant analytical expression is Eq.~\eqref{eq_app_nB} in \ref{sec_app_C}).

\begin{figure}[bhtp!]
		\centering
		\includegraphics[width=1\linewidth]{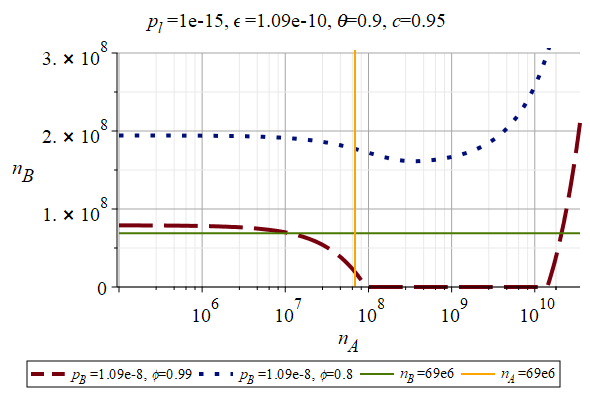}
		\caption{Fatality-free miles that need to be driven in City-B (by Version-B), given that $n_A$ fatality-free miles have been driven in City-A (by Version-A) in scenarios Q4 and Q5.
		The straight horizontal and vertical lines show the amount of road testing that would yield the target confidence $c=95\%$ in the required bound $\mathit{pfm}\leqslant {1.09}e{-8}$ in the single-version, single-city scenario of Q1.}
		\label{fig_nA_vs_nB_two_d}
	\end{figure}

	Fig.~\ref{fig_nA_vs_nB_two_d} shows the answers to Q4 and Q5 as a function of $n_A$, given the same prior knowledge in Q1 (i.e. $p_l=1e{-15}$, $\epsilon={1.09}e{-10}$ and $\theta=0.9$) with $\phi$ prior confidence that the \textit{pfm} of City-B/Version-B is no worse than the \textit{pfm} of City-A/Version-A. For $\phi=0.8$, there are 2 stages on the dotted curve -- it first decreases and then increases ($n_B \to \infty$ as $n_A \to \infty$, and the growth of $n_B$ is $O(n_A)$, as proved in \ref{sec_app_D}), with a global minimum point (cf. Eq.~\eqref{eq_app_nA} in \ref{sec_app_C} for analytical results):
	\begin{itemize}
		\item in the first stage, the more fatality-free miles $n_A$ are observed for City-A/Version-A, the more we believe both cities/versions are safe, thanks to the prior knowledge about ``B is no worse than A''. That is, on the prior distribution in Fig.~\ref{fig_worst_case_joint_prior}, increasing $n_A$, the miles driven, gradually 
		depletes probability mass $M_3$ to the benefit of masses $M_5$ (i.e., probability of the required bound being satisfied for B) and $M_1$. Thus, claiming the required safety level requires less fatality-free evidence to be collected for City-B/Version-B.
		\item the second stage, on the other hand, represents a ``too good to be true'' case. When we increase $n_A$ to a very large value, starting from the prior distribution in Fig.~\ref{fig_worst_case_joint_prior}, both probability masses $M_3$ and $M_5$ gradually decrease while $M_1$ increases (the small prior doubt that $\mathit{pfm}_A$ may be worse than $\mathit{pfm}_B$ becomes very large). But, the probability mass $M_1$ is the probability of \emph{not} satisfying the required bound $p_B$. To show that City-B/Version-B is indeed as safe as required ($Y\leqslant{p_B}$ with probability 95\%)), we need to drive so many miles in B that enough probability mass ``flows' back from  $M_1$ to  $M_5$.
	\end{itemize}
	
	Similarly for the dashed curve in Fig.~\ref{fig_nA_vs_nB_two_d} when $\phi=0.99$, in addition to the two stages discussed above, there is a $n_B=0$ ``terrace'' stage between them -- for this range of $n_A$ values, there is no need to test City-B/Version-B, since the $n_A$ evidence from City-A/Version-A, together with the $\phi$ confidence in B being safer than A, is already enough to prove the claim for B.
	The shape of the curve is the same as for $\phi=0.8$, but truncated at zero in the range of $n_A$ where the required confidence in the bound $p_B$ is exceeded, without any testing in B.

	It is worth mentioning the special case in which we are \textit{certain} the $\mathit{pfm}_B$ is no worse than $\mathit{pfm}_A$ (i.e. $\phi=1$). Then, the result \eqref{eq_worst_case_2d_posterior} becomes
	\begin{equation}
	\label{eq_2d_posterior_phi_equal_1}
	\frac{(1-\epsilon)^{n_A+n_B}\theta}
	{(1-\epsilon)^{n_A+n_B}\theta+(1-p_B)^{n_A+n_B}(1-\theta)} 	
	\end{equation}
	which coincides, according to Theorem \ref{thrm_1}, with the posterior confidence on a required bound $p_B$ after seeing no fatality in $n_A+n_B$ miles (i.e. $Pr(\mathit{pfm}_B \leqslant p_B\mid k=0, n=(n_A+n_B))$). That is, if $\phi=1$, the fatality-free evidence about City-A/Version-A can be treated as evidence about City-B/Version-B as well. This is not the case when $\phi<1$. For example, to claim $\mathit{pfm}_B\leqslant 1.09e{-8}$ with 95\% confidence: (i) if $\phi=1$ and $n_A$ is around 69 million miles, then we don't need any further road testing in City-B/Version-B; But (ii) if $\phi=0.99$ and $n_A$ is still 69 million miles, then we need fatality-free miles $n_B$, around 19 million miles (the intersection point of the dashed curve and the vertical line in Fig.~\ref{fig_nA_vs_nB_two_d}), to ``compensate'' for that $0.01$ doubt.
	
	Now (iii) if $\phi=0.8$ and $n_A$ is still 69 million miles, then we need $n_B$ to be around 170 million miles (the intersection point of the dotted curve and the vertical line in Fig.~\ref{fig_nA_vs_nB_two_d}). This 3rd case reveals an apparent paradox: to conservatively claim that the AV is 100 times safer than humans (with 95\% confidence), City-B/Version-B needs significantly more testing (i.e.~$170$ million miles) than would be needed to conservatively make the same claim for City-A/Version-A (i.e. $69$ million miles) -- this, despite already having driven $69$ million miles for City-A/Version-A and the ``seemingly favourable'' confidence $\phi=0.8$ that City-B/Version-B is safer. An assessor may ask: ``Why then don't I just assess B `from scratch', discarding the evidence of $n_A$ miles driven in A, and $\phi$?". To this one can reply that: (a) discarding knowledge that one has may be unwise, although comparing the results one obtains from using this knowledge with those obtained without it may be informative; (b) the assessment of City-A/Version-A benefits from strong prior knowledge/beliefs in the engineering goal ($\mathit{pfm}_A \leqslant\epsilon$) being achieved, with $\theta$ confidence. However, for City-B/Version-B, such confidence in the engineering goal being met is replaced by the weaker -- though still helpful -- premise that B enjoys greater safety than A with probability $\phi$.
	
	
	\section{Potential fallacies in attempts at ``conservatism''}
	\label{sec_fallacy_be_conservative}
	
	The idea that for certain safety-related decisions one would want ``conservative'' assessments -- to make sure to avoid errors in the direction of excessive optimism even if this causes some error in the direction of pessimism -- is quite commonly accepted.
	However, it is worth pointing out that how to obtain ``conservatism'' is not intuitively obvious.
	The apparatus of theorems that accompany each CBI method is thus necessary.	In particular, which detailed prior gives the most conservative conclusion, given the prior knowledge actually available and the new observations, \emph{depends on the objective function} that we seek to maximise or minimise.
	Despite this being well known, a likely error seems to be that of taking assumptions that are conservative from one viewpoint (i.e. for a certain objective function) and trusting that they ensure conservatism from every viewpoint (see e.g.  \cite[Sec. 7.5]{strigini_bounds_2014}).
	
	We illustrate here the degree of error that this misunderstanding may cause in the scenarios considered in this paper.
	For purposes of comparison with the RAND study, we have introduced CBI theorems to produce worst cases for a specific objective function: a posterior confidence in a required bound on failure rate.
	One may be interested in other reliability measures. For instance, to cite some considered in previous CBI studies, \emph{expected failure rate}, or \emph{probability of suffering no failures in operation}. 
	But the means for being conservative must vary depending on the objective function:
	 there is no universal ``worst-case prior distribution'' for all objective functions of possible interest. 
	Thus, even if we start with the same prior knowledge, the worst-case prior distribution may vary depending on the objective functions chosen. 
	
Misuse of worst-case prior distributions -- i.e., using the prior that is ``worst-case'' for one objective function in order to obtain a worst case for another objective function -- will produce misleading results, with any \emph{errors being in the direction of  unjustified optimism}. 
Shown below are some examples of such misuse. 
	
\begin{figure}[h!]
		\centering
		\includegraphics[width=1\linewidth]{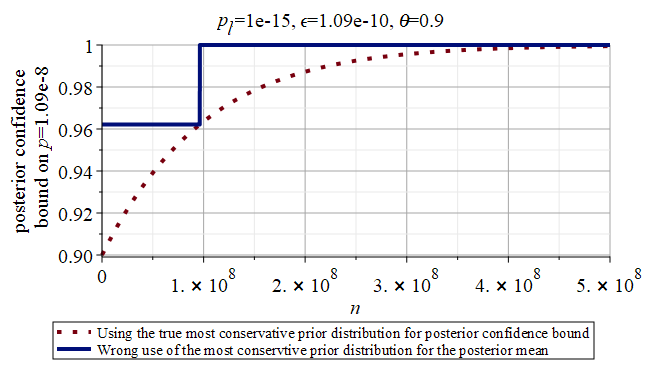}
		\caption{An example of using the wrong worst-case prior distribution.}
		\label{fig_wrong_prior_1}
	\end{figure}

\begin{figure}[h!]
		\centering
		\includegraphics[width=1\linewidth]{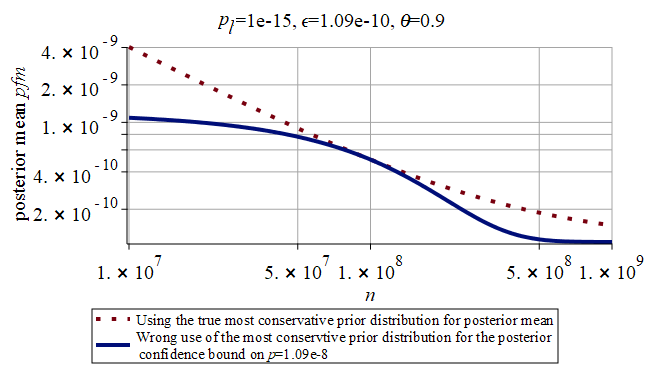}
		\caption{Another example of using the wrong worst-case prior distribution.}
		\label{fig_wrong_prior_2}
	\end{figure}

	In a similar context to that of Q1 (i.e. observing $n$ fatality-free miles and given the partial prior knowledge of Eq.~\eqref{eq_prior_constraints_1}), for an objective function ``posterior expected \textit{pfm}'', a previously proven CBI theorem \cite{bishop_toward_2011} guarantees that:
	
\begin{align}
\MyExp[X\,\vert\,n\mbox{ fatality-free miles}]& =\frac{\int_{p_l}^1 x(1-x)^n \diff F(x)}{\int_{p_l}^1 (1-x)^n \diff F(x)}  \nonumber\\
&\leqslant \frac{\epsilon(1-\epsilon)^n \theta+q(1-q)^n (1-\theta)}
{(1-\epsilon)^n \theta+(1-q)^n (1-\theta)}
\label{eq_CBI_tse2011_illust}
\end{align}
which implies the worst-case prior distribution is still a two-point one: $Pr(X=x)=\theta{\bf 1}_{x=\epsilon} + (1-\theta){\bf 1}_{x=q}$ where the r.h.s. point $q$ is 
a function of $n$ that can be obtained by numerical optimization.
This worst-case prior distribution is different from the one for Q1, in which a posterior confidence in a given confidence bound is of interest -- the left-hand distribution in Fig.~\ref{fig_two_point_priors} which has a fixed far-end point at $x=p$ (where $p$ is the required bound). If we now misuse the worst-case prior related to the posterior expected value Eq.~\eqref{eq_CBI_tse2011_illust}, by applying it to question Q1 in a naive attempt at obtaining worst-case confidence in a bound $p$, this will lead to optimistic results, as Fig.~\ref{fig_wrong_prior_1} illustrates. Or, the other way around, Fig.~\ref{fig_wrong_prior_2} shows an example of using the wrong prior to calculate the worst-case posterior expected \textit{pfm}, which also ends up being optimistic. The intersection points of the curves in both figures represent the special case of some observed $n$ such that, on the worst-case 2-point prior distribution yielding \eqref{eq_CBI_tse2011_illust}, the r.h.s. optimised point $q$ happens to be equal to the given $p$.
	
	\section{Potential fallacies in using disengagement data and/or extrapolating past trends for safety assessment}
	\label{sec_waring_dpm_srgm}
	
	For the sake of public safety, it is usually required that AVs being tested on public roads be supervised by human drivers (sometimes called ``safety drivers'') who are responsible for monitoring the safe operation of the vehicles at all times, and must take over  control (this is a ``disengagement'' of the autonomous driving function) in the event of a failure of the autonomous technology or other emergency. It is also mandated that disengagements during road testing be reported, and records made available
	to the public\footnote{E.g., \url{www.dmv.ca.gov/portal/dmv/detail/vr/autonomous/testing/} (at the time of writing).}.

	As AV technology evolves, one would expect a decreasing trend in the frequency of disengagements. Indeed, Banerjee and co-authors, using large-scale AV road testing data, show negative correlation between \textit{dpm} and cumulative miles driven over three years, but still not reaching AV manufacturers' targets despite millions of miles driven  \cite{banerjee_hands_2018}. Disengagements are much more frequent than serious accidents. So, studying the trend of \textit{dpm}, as done in several statistical papers \cite{banerjee_hands_2018,favaro_examining_2017,dixit_autonomous_2016}, is appealing.
	
	Studying these trends is a useful tool for planning future road testing. To this end, in our previous paper \cite{zhao_assessing_2019} we showed how past AV disengagement data can be used to predict future disengagement via \emph{Software Reliability Growth Models (SRGMs)} \cite{Miller1986EOS} -- a family of statistical, point-process models representing the occurrence of software failures in continuous time. Fitting these models to Waymo's publicly available disengagement data over 51 months, we evaluated the accuracy of their reliability forecasts, and showed how the models' predictions can be improved by ``recalibration'' -- a model improvement technique that utilizes statistical data on how the models' past predictions fall short of observed outcomes \cite{brocklehurst_techniques_1996}.
	
	SRGMs, together with the forecast accuracy evaluation and recalibration methods introduced by \cite{brocklehurst_recalibrating_1990,brocklehurst_techniques_1996}, have been shown to be a powerful tool for extrapolating the trend of AV disengagements in \cite{zhao_assessing_2019}. However, we argued briefly in \cite{zhao_assessing_2019} that SRGMs are \textit{not} suitable for deciding whether the AV is safe enough, even if a SRGM's forecast accuracy and calibration properties have been proved good for a data series. We give here some more detailed and general arguments against using  statistical predictors of \textit{dpm} as indicators of trends in safety (as measured e.g. by \textit{pfm}).

	
	It is true that many disengagements represent safety-relevant events -- ``near misses'' -- and studying near misses is a powerful aid for learning about the safety of a system: to gauge how far from the safety target a system is that has not yet had any accidents, and to correct flaws before they can cause accidents. However:

	\textit{a) The disengagement rate is not a safety indicator}. Generally, there are two types of disengagement -- passive disengagement and active disengagement. Passive disengagements are when subsystems in the AV detect a failure of the autonomous technology, or detect a dangerous situation, and initiate the disengagement. Active disengagements are when the AV does not detect any problem, but the driver monitoring the situation actively disengages the autonomous mode \cite{lv_analysis_2018}. In both cases, ensuring safety relies on the human driver's supervision ability -- the ability to react quickly and correctly. Imperfect human supervision causes noise in the disengagement data -- both false positives and false negatives. Example false positives include unnecessary interventions due to a lack of confidence of the human driver in the AV's ability to negotiate some new pattern of traffic, or due to non-hazardous ``failures'' like uncomfortable riding. Example false negatives include dangerous situations that are not recognised by either the AV or the human driver, but bring no noticeable consequence due to ``good luck''. Moreover, the human supervision ability will vary in complex ways as the autonomy capacity evolves over time \cite{koopman_safety_2019}, which makes it harder to filter out for safety assessment this noise in the disengagement data. Thus, interpreting \textit{dpm} as an indicator of AV safety is wrong \cite{banerjee_hands_2018} and potentially dangerous, through both being misleading and creating incentives to improve \textit{dpm} rather than safety.
	
	Proper use of disengagement data in arguing safety would require assessing the interplay between (i) the evolution of ML functions, (ii) that of the safety drivers' supervision ability, and (iii) the safety subsystems. For instance, we need to consider that an improvement of ML-based functions most likely reduces drivers' ability to trigger disengagements when needed.
	Indeed, it is likely to affect their situation awareness (and thus their ability to detect potentially dangerous situations) and/or their trust in the AV (and thus their readiness to believe that their intervention is needed to resolve that situation).
	Also, the probability of a safety subsystem (like a human driver) taking successful action depends on the probability distribution of the demands on it created by the ML-based functions' failures \cite{SorkinWoods_humanMonitors1985,mammographyDSN2003,PopovStrigini2010ISSRE}, which will vary as the ML-based system evolves.

	\textit{b) No guarantee of monotonicity}.
	Some statistical properties of AVs, such as \textit{dpm}, exhibit a reliability growth trend, in a  statistical sense. However, there is a general reason for not relying on detected trends for safety assessment -- a growth trend  does not imply that \emph{every} update is an improvement; some updates can \emph{reduce} reliability. This is true for all systems, and more so for AVs with their substantial ML components. Thus any statistical predictor like SRGMs, that extrapolates such a trend, will be untrustworthy. That is, we cannot give high confidence in the one prediction that matters for current safety, i.e. the prediction made after the \textit{latest} change. That change could have departed from the previous trend -- even radically increasing the failure rate -- but the predictor would not ``notice'', until the next failure occurs, surprisingly early in view of that last prediction. 
	
	To support the use of 
	trend extrapolation techniques for safety, a possible sufficient condition could be a sound empirical argument (yet to be invented, perhaps infeasible), that the AV is evolving in such a way that it indeed becomes more and more reliable (in terms of hazardous failures) with \emph{every} update. This requires more understanding of the online learning mechanism of the ML component and how the \textit{dynamic} distribution of failures created by it interacts with the safety subsystems. Perhaps such an argument could be supported if the training process for the ML were restricted in some appropriate way. All this relates to activities that accumulate confidence in ``no worse than existing systems'' arguments \cite{littlewood_reliability_2019} (which we also used to answer Q4 and Q5 in Section \ref{sec_CBI_2d}). For example, accelerated testing in simulators before the release of a new version, attempts to build libraries of regression tests, and verification on the learning with safety constraints \cite{pathak_verification_2018}.

	
One may wish to, instead, use an SRGM as a way of obtaining a prior belief to plug into a CBI argument. Indeed this would be quite an appropriate use of extrapolation from previous trends. Care would be needed to choose SRGMs that produce predictions in terms of failure rates, or to rigorously translate predictions that are in terms of distribution of time-to-failure into distributions of failure rates. Apart from this, we see two kinds of practical difficulties that need to be overcome:
\begin{itemize}
    \item obtaining a believable data series as input to the SRGM. One would typically look for a prediction of the frequency of hazardous situations due to AV behaviour. Predicting an accident rate directly would seem inappropriate anyway, and the log of accidents would be too sparse to give a basis for the SRGM to ``learn'' a trend; logs of disengagements would need to be ``cleaned'' from all the spurious events (or missing events: undetected hazardous situations) discussed in item (\textit{a}) earlier on. However, targeted improvements in monitoring and log analysis could possibly overcome these difficulties.  
    \item our CBI arguments have used strong beliefs in some very small upper bound on the failure rate. To use an SRGM's prediction as this belief, one would need the SRGM to actually produce such an output, and to have proved well-calibrated regarding that level of belief: e.g., for its ``99\% confidence'' statements to have been generally correct (or conservative, in the sense we use here for ``conservatism'': conservative in the error they produce on posterior probabilities) for a series of data points. SRGMs have been rarely evaluated for good calibration for these high confidence levels. However, this may be worth doing, including dealing with the complexities in translating between predictions of rates and of time to next event (only the latter being observable). 
\end{itemize}

	\section{Related work}
	\label{sec_related_work}
	
	CBI was initially presented for assessing the reliability of conventional SCSs in \cite{bishop_toward_2011}. Several extensions, e.g., \cite{strigini_software_2013,zhao_modeling_2017,zhao_conservative_2015,zhao_conservative_2018}, have been developed, considering different prior knowledge and objective functions. CBI has recently been used for estimating catastrophic failure related parameters in the runtime formal verification of robots \cite{zhao_probabilistic_2019}, and forming safety arguments in assurance cases for deep learning \cite{zhao_safe_2020}.

	Similarly to \cite{littlewood_reliability_2019}, we have extended CBI from previous ``single system/environment'' applications to a scenario in which multiple systems/environments need to be considered. But the bivariate CBI theorem presented here in Section \ref{sec_CBI_2d} is novel, since: i) the reliability measure of interest in \cite{littlewood_reliability_2019} is a posterior expected failure rate, while here we consider a posterior confidence bound; and ii) the evidence being observed in \cite{littlewood_reliability_2019} is only from the new system/environment, while in this paper we model evidence generated from both systems/environments (i.e. a more general form of the likelihood function).

	
	
	Studies in \cite{banerjee_hands_2018,lv_analysis_2018,favaro_examining_2017,favaro_autonomous_2018,dixit_autonomous_2016} provide descriptive statistics on AV safety and reliability. Both \cite{kalra_driving_2016} and \cite{koopman_autonomous_2017}
	conclude that road testing alone is inadequate evidence of AV safety, and argue the need for other methods to supplement testing on public roads.
	We agree, and our CBI approach provides a concrete way to incorporate such essential prior knowledge into the assessment. 
	There are preliminary Bayesian frameworks, e.g., \cite{cukic_bayesian_2000,smidts_software_2002}, trying to account for verification and validation activities performed prior to testing. However, they assume parametric families (e.g., Beta) for prior distributions, while our CBI approach does not have this restriction and is thus more practical, and more trustworthy (through not demanding unrealistically detailed input and guaranteeing conservatism) for decisions on safety.

	\section{Conclusions \& future work}
	\label{sec_conclusions}
	
	The use of ML solutions in safety-critical applications is on the rise. This imposes new challenges on safety and reliability assessment.
	For ML systems, the inability to directly verify that a design matches its requirements, by reference to the process of deriving the former from the latter \cite{bloomfield_disruptive_2019,koopman_credible_2019}, makes it even harder (compared to conventional software) to estimate the probabilities of failures  \cite{johnson_increasing_2018}. Thus, we believe, increased reliance on operational testing to study failure probabilities and consequences is inevitable, which may form important evidence in heterogeneous safety arguments for autonomous systems \cite{koopman_credible_2019}.
	
	In the case of AVs, the problem is also one of demonstrating ``ultra-high reliability'' \cite{littlewood_validation_1993}, for which it is well known that convincing arguments based on operational testing \textit{alone} are infeasible. While Bayesian inference supports combining operational testing with other forms of evidence, this latter evidence would need to be such as to support very strong prior beliefs. Use of safety subsystems -- not relying on the AV's core ML-based systems -- that are verifiable with conventional methods so as to support stronger prior beliefs (than can be had for the ML-based primary system), would provide part of the solution. How to support prior beliefs strong enough to give sufficient posterior confidence in the kind of dependability levels now desired for AVs requires investigation \cite{littlewood_validation_2011}.
	
	Our CBI approach removes the other major difficulty with these problems, that of trusting more detailed prior beliefs than the evidence typically allows one to argue. One can, thus, take advantage of Bayesian combination of evidence 
	while avoiding optimistic bias (which we found in some other statistical inference models). 
	
	CBI is not limited to ultra-high reliability and certainly does not solve all of the problems of assessing ultra-high reliability, but it does allow one to trust the statistical inference step itself. While it will help to detect some possible flaws in arguments for ultra-high reliability, it will deliver enough confidence when reliability requirements are not so extreme (cf. Fig.~\ref{fig_less_stringent_claim}).

	We demonstrate CBI on one of the most visible examples of ML-based systems with safety assessment challenges -- autonomous vehicles. To recap, the main contributions of this paper are:
	
	\textit{a) for the assessment of constant event rates}, we propose a new variant of the CBI method as a constant event-rate model. This approach will be most useful when there are sound bases for prior beliefs, e.g., through safety-oriented architectures in which the ML-based system functions are paired with non-ML safety subsystems, where such safety subsystems are sufficient to avoid accidents and can be rigorously verified.
	
	Being a Bayesian approach, CBI allows one to ``give credit'' for this essential evidence. It can thus contribute to overcoming the challenges of supporting extreme reliability claims; while its conservatism avoids the potential for dangerous errors in the direction of optimism, inherent in common shortcuts for applying Bayes in these cases.
	
	
	\textit{b) for the assessment of changing failure rates}, as a first step we invent 
	a new bivariate-CBI model utilising prior knowledge on the relationship between the unknown failure rates of before and after the changes. This approach formalises long-established forms of safety arguments about a new system being e.g. ``substantially equivalent'' (for medical devices) or ``globally at least equivalent'' (in the European railway sector) to an earlier system \cite{littlewood_reliability_2019}, which we also have seen being used informally for AVs. 
	
	Using bivariate CBI, we show what various levels of confidence in the relationship between the two failure rates, and the amount of new testing evidence needed, allow one to claim about the reliability of a new system version, or the same system in a new environment.
	
	\textit{c) Warnings against fallacies about ``conservatism''}. The CBI approach, either used with a constant or changing event-rate model, has the unique advantage of being conservative (that is, of avoiding errors in the direction of optimism) in assessing SCSs.
	While it is known that ``being conservative'' depends on the objective function chosen, trying to achieve conservatism without the necessary mathematical proofs is known to produce fallacies. We discuss these possible fallacies, with illustrative examples.
	
	\textit{d) Warnings against using disengagement data, and/or extrapolation of observed trends, for safety assessment}. Disengagements are the most widely reported statistical data about AVs. There is an obvious decreasing trend of \textit{dpm} as the AV technology matures over time. Predicting this trend via statistical tools, e.g., SRGMs in \cite{zhao_assessing_2019}, is feasible, and useful for non-safety decisions. For instance, the trend of disengagement data is used to gauge the ``stability/maturity'' of AVs in \cite{banerjee_hands_2018}. In the present paper, we discuss why reliance on disengagement data, and/or on trend extrapolation, for safety assessment may be dangerous.
	

	In future work, we plan to:
	
\textit{(i)} explore practical means for quantitatively stating prior knowledge (e.g. evidence from various system verification methods, applied to the AV system and its subsystems) for input to our CBI method, detailing the processes that we outlined in Section \ref{sec_CBI}; 
	
	\textit{(ii)} adapt CBI extensions to base decisions directly on risk of accidents/fatality-free operation over finite periods, instead of focusing on a specific bound on failure rate. As argued elsewhere  \cite{strigini_software_2013}, this would more directly support sound decisions about the progressive introduction of AVs;
	
	\textit{(iii)} build more detailed safety arguments for architectures using safety subsystems, with appropriate subsystem-level arguments based on the different forms of evidence available about the various subsystems. These arguments would be more easily adapted to evolving ML subsystems;
	
	\textit{(iv)} represent plausible forms of failure correlation (over successive miles driven) within the statistical model in our CBI approach. As outlined in section \ref{sec_OT_and_failure_process}, our present CBI model assumes that the fundamental AV failure process is Bernoulli -- specifically, that failures over successive miles driven are \emph{statistically independent and identically distributed} (i.i.d) in their occurrence. However, there are a number of analogous assessment scenarios where such i.i.d. assumptions may only hold very approximately, if at all \cite{strigini_testing_1996,goseva-popstojanova_failure_2000,Trivedi_1993}. In future work, upon explicitly incorporating failure correlation into CBI models, we will quantify the extent to which the i.i.d. assumption may undermine conservative assessments. 
	
	
	Focusing on the ``hot'' area of AVs, and the ``ultra-high reliability'' problems that they pose, inevitably led us to highlight remaining problems and extensive work still necessary. However, the novel CBI theorems that we have presented are generally applicable. They are a useful tool for ameliorating the problem of assessing AVs, and for solving many current assessment problems. The main contribution of CBI is to free users of Bayesian methods from the risk of inordinately optimistic predictions, which arise from spurious prior assumptions introduced for mathematical convenience. Our numerical examples show that this guaranteed conservatism does not necessarily lead to excessively pessimistic predictions.
	Even in cases where CBI yields disappointing conclusions -- the desired claims are not supported -- CBI helps assurance: (i) it encourages clarity about how the evidence collected translates into logical arguments; (ii) it reveals gaps between the evidence brought to the argument and the claims one wishes to support; and thus (iii) helps to orient design and verification towards producing appropriate evidence.

	\appendix
	
	\section{Statement and proof of CBI Theorem \ref{thrm_1}}
	\label{sec_app_A}
	\noindent \textbf{Problem}: 
	Consider the set ${\mathcal D}$ of all probability distributions defined over the unit interval, each distribution representing a potential prior distribution of \emph{pfm} values for an AV. For $0<p_l<\epsilon\leqslant 1$,
	we seek a prior distribution that minimises the posterior confidence in a reliability bound $p\in[p_l, 1]$, given $k$ fatalities have occurred over $n$ miles driven and subject to constraints on some quantiles of the prior distribution. That is, for $\theta\in(0,1]$, we solve 
	\begin{equation*}
	\begin{aligned}
	& \underset{{\mathcal D}}{\text{minimise}}
	& & Pr(X \leqslant p\mid k\&n) \\
	& \text{subject to}
	& & Pr(X \leqslant \epsilon)=\theta,\quad Pr(X\geqslant p_l)=1
	\end{aligned}
	\end{equation*}
	
	\noindent \textbf{Solution}: There is a prior in ${\mathcal D}$ that gives the infimum for the posterior confidence: the 2-point distribution $$Pr(X=x)=\theta{\bf 1}_{x=x_1} + (1-\theta){\bf 1}_{x=x_3}$$ where $p_l\leqslant x_1 \leqslant \epsilon < x_3\,$, and the values of $x_1$, $x_3$ both depend on the model parameters (i.e. $p_l, \epsilon, p$) as well as $k$ and $n$. Using this prior, the infimum for the posterior confidence is
	\begin{align}
	\frac{x_1^k(1-x_1)^{n-k}\theta}{x_1^k(1-x_1)^{n-k}\theta + x_3^k(1-x_3)^{n-k}(1-\theta)}{\bf 1}_{p>\epsilon}
	\label{eq_res_CBI_post_conf_bound_see_failures_app}
	\end{align}
	where ${\bf 1}_{\tt S}$ is an indicator function -- it is equal to 1 when {\tt S} is true and 0 otherwise.\newline

	\begin{proof} The proof is constructive, starting with \emph{any} feasible prior distribution and progressing in 3 stages, each stage producing priors that give progressively worse posterior confidence than in the previous stage. In more detail, assuming $\epsilon \leqslant p$ (the argument for $p<\epsilon$ is analogous):
		\begin{enumerate}
			\item First we show that, for any given feasible prior distribution in $\mathcal D$, there is an  equivalent feasible 3-point prior distribution. ``Equivalent'', in that the 3-point distribution has the same value for the posterior confidence in $p$ as the given feasible prior. Consequently, we restrict the optimisation to the set ${\mathcal D}^\ast$ of all such 3-point distributions;
			\item For each prior in ${\mathcal D}^\ast$, there exists a 2-point prior distribution with a smaller posterior confidence in $p$. Consequently, we restrict the optimisation to the set ${\mathcal D}^{\ast\ast}$ of all such 2-point priors;
			\item A monotonicity argument determines a 2-point prior in ${\mathcal D}^{\ast\ast}$ with the smallest posterior confidence in $p$. 
		\end{enumerate}
		
		\emph{Stage 1}: Assuming $\epsilon \leqslant p$, note that for any prior distribution $F\in{\mathcal D}$, we may write
		\begin{align}
		\label{eq_appendix_ob}
		Pr(X \leqslant p\mid k\&n)=\frac{T }{T+\int_{p^+}^{1} x^k(1-x)^{n-k} \mathrm dF(x)}
		\end{align}
		where $T\!=\!\int_{p_l}^{\epsilon} x^k(1\!-\!x)^{n-k} \mathrm dF(x)+\!\int_{\epsilon^+}^{p} x^k(1\!-\!x)^{n-k} \mathrm dF(x)$. The \emph{mean-value-theorem for integrals} ensures that three points exist, $x_1\in[p_l,\epsilon]$, $x_2\in(\epsilon,p]$ and $x_3\in (p,1]$, such that \eqref{eq_appendix_ob} becomes (denote $\int_{\epsilon^+}^{p}\mathrm dF(x)=\beta$):
		\begin{align}
		\label{eq_appendix_ob2}
		\frac{x_1^k(1-x_1)^{n-k}\theta+x_2^k(1-x_2)^{n-k}\beta }
		{x_1^k(1\!-\!x_1)^{n\!-\!k}\theta\!+\!x_2^k(1\!-\!x_2)^{n\!-\!k}\beta\!+\!x_3^k(1\!-\!x_3)^{n\!-\!k}(1\!-\!\theta\!-\!\beta)}
		\end{align}

		By establishing \eqref{eq_appendix_ob2} we have established that, for \emph{any} given prior distribution one might start off with, there exists an equivalent 3-point prior distribution. Thus, we restrict the optimisation to ${\mathcal D}^\ast$, the set of all of these equivalent priors. 
		
		\emph{Stage 2:} Next, for each prior in ${\mathcal D}^\ast$, there is a 2-point prior distribution that is guaranteed to give a smaller posterior confidence in $p$. To see this for any given prior in ${\mathcal D}^\ast$ with posterior \eqref{eq_appendix_ob2}, treat all of the other variables as fixed (i.e. the ``$x$''s and $\theta$) and consider which of the allowed values for $\beta$, given these fixed values of the other variables, guarantees a distribution that reduces the posterior confidence. The continuous differentiability of rational functions -- of which \eqref{eq_appendix_ob2} is one -- allows the partial derivative of \eqref{eq_appendix_ob2} w.r.t. $\beta$ to show us the way to do this. The partial derivative of \eqref{eq_appendix_ob2} with respect to $\beta$ is always positive, irrespective of the fixed values the $x_i$s take in their respective ranges. So, to minimise \eqref{eq_appendix_ob2}, we set $\beta=0$. This gives the attainable lower bound \eqref{eq_CBI_post_conf_bound_see_failures_app}, attained by the 2-point prior distribution with probability masses $\theta$ at $x=x_1$, and $1-\theta$ at $x=x_3$. Therefore, we restrict the optimisation to ${\mathcal D}^{\ast\ast}$ -- the set of all such priors.
		\begin{align}
		Pr(\!X \!\leqslant\! p \mid k\&n) &\geqslant  \frac{x_1^k(1-x_1)^{n-k}\theta }{x_1^k(1-x_1)^{n-k}\theta\!+\!x_3^k(1-x_3)^{n-k}(1-\theta)}\nonumber \\
		&= \frac{1}{1 + \left(\frac{x_3^k(1-x_3)^{n-k}}{x_1^k(1-x_1)^{n-k}}\right)\frac{1-\theta}{\theta}}
		\label{eq_CBI_post_conf_bound_see_failures_app}
		\end{align}
		
		\emph{Stage 3:} To minimise \eqref{eq_CBI_post_conf_bound_see_failures_app} further (and, thereby, obtain optimal priors in ${\mathcal D}^{\ast\ast}$), we maximise $x_3^k(1-x_3)^{n-k}$ and minimise $x_1^k(1-x_1)^{n-k}$ over the allowed ranges for $x_1,x_3$. The problem is now reduced to a simple monotonicity analysis given different values of the other model parameters, as follows. Since $x^k(1-x)^{n-k}$ is bell-shaped over $[0,1]$ with a maximum at $x=k/n$, the following defines 2-point priors that solve the optimisation problem (depicted in Fig~\ref{fig_all_priors}):
		
		\begin{itemize}
			\item When $0\leqslant k/n \leqslant p_l$:
			\subitem to minimise $x_1^k(1-x_1)^{n-k}$, subject to $x_1\in[p_l,\epsilon]$, we set $x_1=\epsilon$;
			\subitem to maximise $x_3^k(1-x_3)^{n-k}$, subject to $x_3\in (p,1]$, we set $x_3=p$.
			\item When $p_l<k/n\leqslant \epsilon$, and $p_l^k(1-p_l)^{n-k} \geqslant \epsilon^k(1-\epsilon)^{n-k}$:
			\subitem to minimise $x_1^k(1-x_1)^{n-k}$, subject to $x_1\in[p_l,\epsilon]$, we set $x_1=\epsilon$;
			\subitem to maximise $x_3^k(1-x_3)^{n-k}$, subject to $x_3\in (p,1]$, we set $x_3=p$.
			\item When $p_l<k/n\leqslant \epsilon$, and $p_l^k(1-p_l)^{n-k} < \epsilon^k(1-\epsilon)^{n-k}$:
			\subitem to minimise $x_1^k(1-x_1)^{n-k}$, subject to $x_1\in[p_l,\epsilon]$, we set $x_1=p_l$;
			\subitem to maximise $x_3^k(1-x_3)^{n-k}$, subject to $x_3\in (p,1]$, we set $x_3=p$.
			\item When $\epsilon < k/n \leqslant p$:
			\subitem to minimise $x_1^k(1-x_1)^{n-k}$, subject to $x_1\in[p_l,\epsilon]$, we set $x_1=p_l$;
			\subitem to maximise $x_3^k(1-x_3)^{n-k}$, subject to $x_3\in (p,1]$, we set $x_3=p$.
			\item When $p < k/n \leqslant 1$:
			\subitem to minimise $x_1^k(1-x_1)^{n-k}$, subject to $x_1\in[p_l,\epsilon]$, we set $x_1=p_l$;
			\subitem to maximise $x_3^k(1-x_3)^{n-k}$, subject to $x_3\in (p,1]$, we set $x_3=k/n$.
		\end{itemize}
		
		\begin{figure*}[ht]
			\centering
			\includegraphics[width=0.8\linewidth]{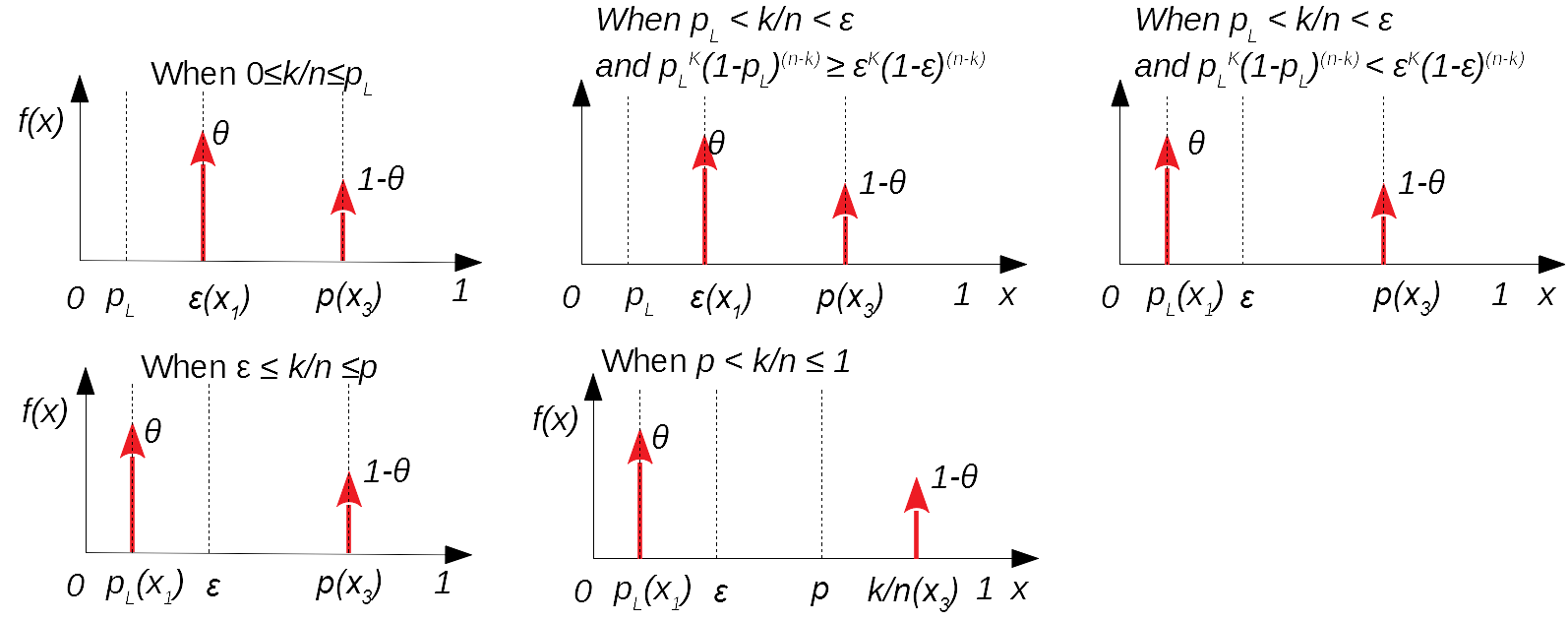}
			\caption{The 5 possible cases of two-point prior distributions that minimise \eqref{eq_appendix_ob}. Notice the important role of where $k/n$ lies.}
			\label{fig_all_priors}
		\end{figure*}
		\noindent The form of $Pr(X < p\mid k\&n)$ for each prior above is \eqref{eq_res_CBI_post_conf_bound_see_failures_app}. All of the preceding arguments guarantee that this value is the infimum for $Pr(X \leqslant p\mid k\&n)$.
		
		We have thus proved Theorem \ref{thrm_1} for $\epsilon\leqslant p$. Let us begin the optimisation again, but now assuming $p<\epsilon$. For any feasible prior $F\in{\mathcal D}$, the objective function $Pr(X \leqslant p\mid k\&n)$ can be written as
		\begin{align}
		\label{eq_appendix_ob3}
		\frac{L}{L+\int_{p^+}^{\epsilon} x^k(1-x)^{n-k} \mathrm dF(x)+\int_{\epsilon^+}^{1} x^k(1-x)^{n-k} \mathrm dF(x)}
		\end{align}
		
		where $L=\int_{p_l}^{p} x^k(1-x)^{n-k} \mathrm dF(x)$. As before, the \emph{mean-value-theorem} ensures the existence of three points $x_1,x_2,x_3$ in the ranges: $x_1\in [p_l,p], x_2\in (p,\epsilon],x_3\in (\epsilon,1]$ such that \eqref{eq_appendix_ob3} becomes (denote $\int_{p_l}^{p}\mathrm dF(x)=\gamma$, where $0 \leqslant \gamma \leqslant \theta$):
		\begin{align}
		\label{eq_appendix_ob4}
		\frac{L'}
		{L'+x_2^k(1-x_2)^{n-k}(\theta-\gamma)+x_3^k(1-x_3)^{n-k}(1-\theta)}
		\end{align}
		where $L'=x_1^k(1-x_1)^{n-k}\gamma$.
		
		The derivative of \eqref{eq_appendix_ob4} with respect to $\gamma$ is always positive, irrespective of the fixed values the $x_i$s can take in their allowed ranges. So, to minimise \eqref{eq_appendix_ob4}, we simply set $\gamma=0$. Thus, \eqref{eq_appendix_ob4} has a lower bound of 0 when $p<\epsilon$, and the corresponding prior distribution that attains this is still a 2-point one with probability masses at $x=x_2$ and $x=x_3$, regardless of what fixed values $x_2$ and $x_3$ take in their allowed ranges.
	\end{proof}
	
	\section{Formal analysis for Q3 in Sec.~\ref{sec_num_examples_CBI} }
	\label{sec_app_B}
	
	We seek to understand what happens when $n_1$ fatality-free driven miles support a \emph{pfm} claim $p$ with confidence $c$. And, upon seeing a fatality after $n_1$ miles, understanding how many more fatality-free miles $n_2$ are needed to maintain support for the claim. So, what follows is an analysis of the asymptotic ``large $n$'' behaviour implied by the worst-case posterior confidence \eqref{eq_CBI_post_conf_bound_see_failures} in Theorem \ref{thrm_1}. Assume $c$ and $\theta$ are given in the practical case when $c\geqslant \theta$.

	Let $n^\ast$ denote the number of miles that satisfies $\epsilon(1-\epsilon)^{n^\ast-1} = p_l(1-p_l)^{n^\ast-1}$. So, from \ref{sec_app_A} above, for $n< n^\ast$ we have $x_1=p_l$, and for $n\geqslant n^\ast$ we have $x_1=\epsilon$. Note that $n^\ast$ is independent of $c$ and $\theta$, so this number of miles will be the same no matter what levels of confidence one is either interested in, or has prior to road testing.

	Now, using \eqref{eq_CBI_post_conf_bound_see_failures}, we may write the number of miles driven as a function of the remaining problem parameters. That is, for $\epsilon<p\leqslant 1$,
	\begin{equation}
	n(c, p, \theta, x_1, k) := k + \left(\frac{k\log(x_1/p) + \log(\frac{\theta(1-c)}{c(1-\theta)})}{\log(\frac{1-p}{1-x_1})}\right)
	\label{eq_n_formula}
	\end{equation}
	where we have assumed that the values of $n$ ensure $k/n\leqslant p$ holds. In particular, for $k=1$, let $p^\ast$ uniquely satisfy
	\begin{equation}
	n^\ast = 1 + \left(\frac{\log(x_1/p^\ast) + \log(\frac{\theta(1-c)}{c(1-\theta)})}{\log(\frac{1-p^\ast}{1-x_1})}\right)
	\label{eq_pstar}
	\end{equation}
	where $x_1=p_l, \epsilon$ both result in the same $n^*$ value, by the definition of $n^\ast$. So, for $p>p^\ast$, we must have $x_1=p_l$. And, for $\epsilon < p\leqslant p^\ast$, we have $x_1=\epsilon$.

	If, for otherwise fixed parameter values, we denote $\tilde{n}$ the number of miles according to \eqref{eq_n_formula} when $k=1$, and $n_1$ the number of miles when $k=0$, then the number of additional miles $n_2$ needed upon seeing a fatality immediately after $n_1$ miles is $n_2 := \tilde{n} - n_1$.
	
	Suppose then, that $p>p^\ast$ and let $p$ tend to $p^\ast$ from above. The following limits follow from the continuity of $n$ in \eqref{eq_n_formula}:
	\begin{enumerate}
		\item \emph{If a fatality is observed (so $k=1$)} then, as $p$ tends to $p^\ast$ from above, we have $x_1 = p_l$, and the number of miles that are needed to be driven to support a claim in $p$ -- with confidence $c$ using prior confidence $\theta$ in the engineering goal $\epsilon$ being met -- is 
		\begin{align*}
		\lim\limits_{p\downarrow p^\ast}\tilde{n}& = \lim\limits_{p\downarrow p^\ast} n(c, p, \theta, p_l, 1)  \\
		&= n(c, \lim\limits_{p\downarrow p^\ast} p, \theta, p_l, 1) = n(c, p^\ast, \theta, p_l, 1) = n^\ast
		\end{align*}
		
		\item \emph{If no fatalities are observed (so $k=0$)} then, as $p$ tends to $p^\ast$ from above, the number of fatality-free miles that are needed to be driven to support a claim in $p$ -- with confidence $c$ using prior confidence $\theta$ in the engineering goal $\epsilon$ being met -- is 
		\begin{align*}
		\lim\limits_{p\downarrow p^\ast}n_1& = \lim\limits_{p\downarrow p^\ast} n(c, p, \theta, \epsilon, 0) =n(c, \lim\limits_{p\downarrow p^\ast} p, \theta, \epsilon, 0) \\
		& =n(c, p^\ast, \theta, \epsilon, 0) = \frac{ \log(\frac{\theta(1-c)}{c(1-\theta)})}{\log(\frac{1-p^\ast}{1-\epsilon})}
		\end{align*}
		\noindent Recall, from \ref{sec_app_A}, that $x_1 = \epsilon$ must hold here for all $p$ when $k=0$.
		
		\item so, using these last two results, the number of extra miles needed is 
		\begin{equation}
		\lim\limits_{p\downarrow p^\ast}n_2 = n^\ast - \frac{ \log(\frac{\theta(1-c)}{c(1-\theta)})}{\log(\frac{1-p^\ast}{1-\epsilon})}
		\label{eq_n2fromabove}
		\end{equation}
	\end{enumerate}
	
	Alternatively, suppose $p<p^\ast$ and let $p$ tend to $\epsilon$ from above. The following limits also follow from \eqref{eq_n_formula}:
	\begin{enumerate}
		\item \emph{If a fatality is observed (so $k=1$)}, then as $p$ tends to $\epsilon$ from above, we have $x_1 = \epsilon$, and the number of miles that are needed to be driven to support a claim in $p$ -- with confidence $c$ using prior confidence $\theta$ in the engineering goal $\epsilon$ being met -- is 
		\begin{equation*}
		\lim\limits_{p\downarrow \epsilon}\tilde{n} = \lim\limits_{p\downarrow \epsilon} n(c, p, \theta, \epsilon, 1) =  n(c, \lim\limits_{p\downarrow \epsilon} p, \theta, \epsilon, 1) = \infty
		\end{equation*}
		
		\item \emph{If no fatalities are observed (so $k=0$)} then, as $p$ tends to $\epsilon$ from above, the number of fatality-free miles that are needed to be driven to support a claim in $p$ -- with confidence $c$ using prior confidence $\theta$ in the engineering goal $\epsilon$ being met -- is 
		\begin{equation*}
		\lim\limits_{p\downarrow \epsilon}n_1 = \lim\limits_{p\downarrow \epsilon} n(c, p, \theta, \epsilon, 0) =  n(c, \lim\limits_{p\downarrow \epsilon} p, \theta, \epsilon, 0) = \infty
		\end{equation*}
		
		\item the last two results show that both $\tilde{n}$ and $n_1$ grow without bound, however the number of extra miles needed is bounded above, since (by \emph{L'Hospital's rule})
		\begin{align}
		\lim\limits_{p\downarrow \epsilon}n_2 &= \lim\limits_{p\downarrow \epsilon}(\tilde{n} -n_1) 	\nonumber\\
		&=\lim\limits_{p\downarrow \epsilon} ( n(c, p, \theta, \epsilon, 1) - n(c, p, \theta, \epsilon, 0) )  \nonumber\\
		& =1 + \lim\limits_{p\downarrow \epsilon}\left(\frac{\log(\epsilon/p)}{\log(\frac{1-p}{1-\epsilon})}\right) \nonumber \\
		&= 1 + \lim\limits_{p\downarrow \epsilon}\frac{(1/p)}{1/(1-p)} = 1 + \frac{1-\epsilon}{\epsilon} = 1/\epsilon
		\label{eq_n2pastpstar}
		\end{align}
		\noindent Note that, like $n^\ast$, this limit is independent of $c$ and $\theta$.
	\end{enumerate}

	\section{Statement and proof of CBI Theorem \ref{theorem_two_d_CBI}}
	\label{sec_app_C}
	\noindent \textbf{Problem}:
	\begin{equation*}
	\begin{aligned}
	& \underset{{\mathcal D}}{\text{minimise}}
	& & Pr(Y\leqslant p_B \mid n_A, n_B) \\
	& \text{subject to}
	& & Pr(X \leqslant \epsilon)=\theta,\quad Pr(X\geqslant p_l)=1\\
	& & & Pr(Y \leqslant X)=\phi,\quad Pr(Y\geqslant p_l)=1
	\end{aligned}
	\end{equation*}
	
	\noindent \textbf{Solution}: There is a three-point prior in ${\mathcal D}$ that gives the infimum for the posterior confidence.
	When $\phi>1-\theta$, as shown in Fig.~\ref{fig_worst_case_joint_prior}, it is $Pr(X=x,Y=y)=(1-\phi){\bf 1}_{x=p_l,y=p_B} + (1-\theta){\bf 1}_{x=p_B,y=p_B}+(\phi-1+\theta){\bf 1}_{x=\epsilon,y=\epsilon}$. Using this prior, the infimum for the posterior confidence is
	\begin{equation}
	\label{eq_AP_worst_case_2d_posterior_}
	\frac{(1-\epsilon)^{n_A+n_B}M_5}
	{(1\!-\!\epsilon)^{n_A+n_B}M_5\!+\!(1\!-\!p_B)^{n_A+n_B}M_3\!+\!(1\!-\!p_l)^{n_A}(1\!-\!p_B)^{n_B}M_1} {\bf 1}_{\phi>1-\theta}	
	\end{equation}
	where $M_1=1-\phi$, $M_3=1-\theta$ and $M_5=\phi-1+\theta$, and again ${\bf 1}_{\tt S}$ is an indicator function -- it is equal to 1 when {\tt S} is true and 0 otherwise. When $\phi \leqslant 1-\theta$, the worst-case prior distribution will always yield 0 as the infimum for $Pr(Y\leqslant p_B \mid n_A, n_B)$. Thus, this case is not of practical interest.	\newline
	
	\begin{proof}
		The proof proceeds in 3 stages:
		\begin{enumerate}
			\item First we show that, for any given feasible prior distribution in $\mathcal D$, there is an  equivalent feasible 7-point prior distribution -- one point for each of the 7 regions in Fig.~\ref{fig_joint_prior_fab}. ``Equivalent'', in that the 7-point distribution yields the same value for the posterior confidence in $\mathit{pfm}_B \leqslant p_B$ as the given feasible prior. Consequently, we restrict the optimisation to the set ${\mathcal D}^\ast$ of all such 7-point distributions;
			\item In ${\mathcal D}^\ast$, for all priors with the same probability mass within each region, we show there is an optimal point within each region that further minimises the objective function. Consequently, we collect all such 7-point priors, with probability masses allocated to these optimal points, as a new set ${\mathcal D}^{\ast\ast}$;
			\item A monotonicity argument determines a 3-point prior (since the other 4 points have 0 probability) in ${\mathcal D}^{\ast\ast}$ that gives the infimum for the posterior confidence in $\mathit{pfm}_B \leqslant p_B$. 
		\end{enumerate}
		
		\textit{Stage 1}: For any prior distribution $F_{AB}(x,y)\in \mathcal D$, by partitioning the sample space into 7 regions as shown in Fig.~\ref{fig_joint_prior_fab}, our objective function of Eq.~\eqref{eq_post_conf_2D} can be rewritten as:
		\begin{align}
		\label{eq_AP_post_conf_2D_OB}
		&Pr(Y\leqslant p_B \mid n_A, n_B)\\ \nonumber
		&=
		\frac{\sum_{i=4}^{i=7}\int\!\!\int_{\text{region}_i} (1-x)^{n_A}(1-y)^{n_B} \mathrm dF_{AB}(x,y) }
		{\sum_{i=1}^{i=7}\int\!\!\int_{\text{region}_i} (1-x)^{n_A}(1-y)^{n_B} \mathrm dF_{AB}(x,y) }
		\end{align}
		The \emph{mean-value-theorem for integrals} ensures that, within each region$_i$, there exits a point $(x_i,y_i)$ such that:
		\begin{equation}
		\int\!\!\int_{\text{region}_i} (1-x)^{n_A}(1-y)^{n_B} \mathrm dF_{AB}(x,y) = (1-x_i)^{n_A}(1-y_i)^{n_B}M_i
		\end{equation}
		Note, $M_i$ is the probability mass associated with region$_i$, and the ranges of $x_i$ and $y_i$ are within the region$_i$. So, the result \eqref{eq_AP_post_conf_2D_OB} becomes:
		\begin{align}
		\label{eq_AP_post_conf_2D_OB_Mi_1}
		Pr(Y\leqslant p_B \mid n_A, n_B)&=
		\frac{\sum_{i=4}^{i=7}(1-x_i)^{n_A}(1-y_i)^{n_B}M_i}
		{\sum_{i=1}^{i=7}(1-x_i)^{n_A}(1-y_i)^{n_B}M_i }
		\end{align}
		By establishing \eqref{eq_AP_post_conf_2D_OB_Mi_1}, we have established that, for \emph{any} given prior distribution one might start off with, there exists an equivalent 7-point prior distribution -- one point for each region$_i$ and with probability mass $M_i$. Thus, we restrict the optimisation to ${\mathcal D}^\ast$, the set of all of these equivalent priors.
		
		\textit{Stage 2}: Slightly rearranging \eqref{eq_AP_post_conf_2D_OB_Mi_1}, we obtain:		
		\begin{align}
		Pr(Y\leqslant p_B \mid n_A, n_B)
		&=\frac{1}{1+\frac{\sum_{i=1}^{i=3}(1-x_i)^{n_A}(1-y_i)^{n_B}M_i}
			{\sum_{i=4}^{i=7}(1-x_i)^{n_A}(1-y_i)^{n_B}M_i }}
		\label{eq_AP_post_conf_2D_OB_Mi_2}
		\end{align}

		Now, for each prior in ${\mathcal D}^\ast$, by fixing the $M_i$s (and assuming they satisfy the constraints), we can ``move'' each point $(x_i,y_i)$ freely within each region$_i$ to further minimise \eqref{eq_AP_post_conf_2D_OB_Mi_2}. 
		
		Since all $M_i \geqslant 0$, to minimise \eqref{eq_AP_post_conf_2D_OB_Mi_2}, we need to minimise the $x_i$s and $y_i$s within region$_1$, region$_2$ and region$_3$, and to maximise the $x_i$s and $y_i$s within region$_4$, region$_5$,  region$_6$ and region$_7$. Note, this observation doesn't depend on the values of the $M_i$s in their range of $[0,1]$. The movement and optimal locations of point masses in each region are depicted in Fig.~\ref{fig_move_points_on_worst_case_joint_prior}, that is:
		\begin{align*}
		&(x_1, y_1) \rightarrow (p_l,p_B), \quad (x_2, y_2) \rightarrow (\epsilon,p_B), \quad (x_3, y_3) \rightarrow (p_B,p_B) \\
		&(x_4, y_4) \rightarrow (\epsilon,p_B), \quad (x_5, y_5) \rightarrow (\epsilon,\epsilon), \quad (x_6, y_6) \rightarrow (p_B,p_B)\\
		& (x_7, y_7) \rightarrow (1,p_B)
		\end{align*}
		
		So, we rewrite the objective function as (note: the term associated with $M_7$ is 0, and thus omitted)
		\begin{align}
		\label{eq_AP_post_conf_2D_OB_after_move}
		&Pr(Y\leqslant p_B \mid n_A, n_B) \nonumber
		\\
		&\geqslant \frac{1}{1+\frac{(1-p_l)^{n_A}(1-p_B)^{n_B}M_1+(1-\epsilon)^{n_A}(1-p_B)^{n_B}M_2+(1-p_B)^{n_A}(1-p_B)^{n_B}M_3}
		{(1-\epsilon)^{n_A}(1-p_B)^{n_B}M_4+(1-\epsilon)^{n_A}(1-\epsilon)^{n_B}M_5+(1-p_B)^{n_A}(1-p_B)^{n_B}M_6}}
		\end{align}
		
		\begin{figure}[bhtp!]
			\centering
			\includegraphics[width=0.7\linewidth]{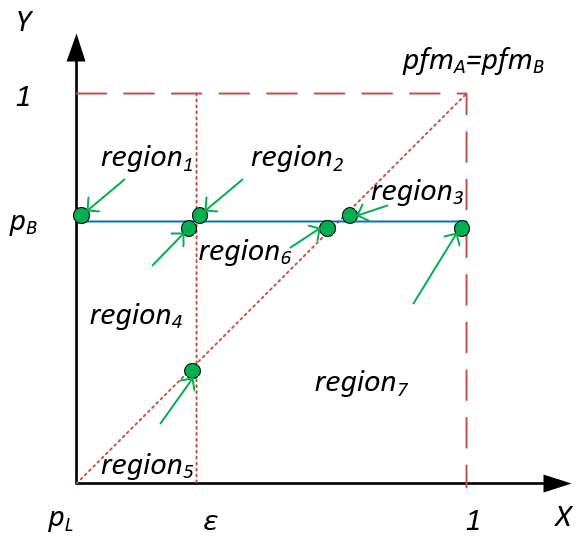}
			\caption{The movement and optimal locations of the point masses in each region, e.g., in region$_5$, the optimal point is at $(\epsilon,\epsilon)$.}
			\label{fig_move_points_on_worst_case_joint_prior}
		\end{figure}
		
		\textit{Stage 3}: Now the problem is reduced to an optimisation problem\footnote{To be exact, it is a linear-fractional programming problem (that may be converted to an equivalent linear programming problem).}
		with an objective function that is the r.h.s. of  Eq.~\eqref{eq_AP_post_conf_2D_OB_after_move}, over the parametric space of the $M_i$s and subject to the constraints:
		\begin{equation}
		\label{eq_AP_prior_constraints_Mi}
		\sum_{i=1,4,5} M_i=\theta, \quad \sum_{i=3,5,7} M_i=\phi, \quad \sum_{i=1}^{i=7} M_i=1
		\end{equation}

		First, we rearrange the three constraints in \eqref{eq_AP_prior_constraints_Mi} and substitute them into the r.h.s. of Eq.~\eqref{eq_AP_post_conf_2D_OB_after_move}. The objective function becomes a function $h$ of $M_3$, $M_4$, $M_5$ and $M_6$:
		\begin{equation}
		\label{eq_AP_post_conf_2D_OB_after_move_2}
		h(M_3,M_4,M_5,M_6) := \frac{1}{1+\frac{Nu(M_3,M_4,M_5,M_6)}{De(M_3,M_4,M_5,M_6)}}
		\end{equation}
		where
		\begin{align}
		&Nu(M_3,M_4,M_5,M_6)=(1-p_l)^{n_A}(1-p_B)^{n_B}(\theta-M_4-M_5) \nonumber\\
		&+(1-\epsilon)^{n_A}(1-p_B)^{n_B}(1-\theta-\phi+M_5-M_6)+(1-p_B)^{n_A+n_B}M_3\\
		&De(M_3,M_4,M_5,M_6)=(1-\epsilon)^{n_A+n_B}M_5+(1-p_B)^{n_A+n_B}M_6 \nonumber\\
		&+(1-\epsilon)^{n_A}(1-p_B)^{n_B}M_4
		\end{align} 
		
		Since we have considered all of the constraints, we may treat $M_3$, $M_4$, $M_5$ and $M_6$ as independent variables. The partial derivative of $h$ in terms of $M_5$ is
		\begin{align}
		\frac{\partial h}{\partial M_5}=\frac{-\frac{\partial Nu}{\partial M_5}De+Nu\frac{\partial De}{\partial M_5}}{(De+Nu)^2}
		\end{align}
		Since, upon taking partial derivatives,
		\begin{align}
		\frac{\partial Nu}{\partial M_5}&=-(1-p_B)^{n_B}\left((1-p_l)^{n_A}-(1-\epsilon)^{n_A}\right) \\
		\frac{\partial De}{\partial M_5}&=(1-\epsilon)^{n_A+n_B}
		\end{align}
		and both $Nu$ and $De$ are positive, we have $\frac{\partial h}{\partial M_5}>0$; this means $h$ is an increasing function of $M_5$.
		
		Similarly, we can prove that $h$ is an increasing function of $M_4$ and $M_6$, and a decreasing function of $M_3$. We omit the proofs for brevity.

		Now, depending on the values of $\theta$ and $\phi$, we have two cases:
		\begin{itemize}
			\item when $\phi \leqslant 1-\theta$, to minimise the objective function $h$, we set $M_5=M_4=M_6=M_7=0$, $M_3=\phi$, $M_1=\theta$ and $M_2=1-\theta-\phi$. In this case, the worst-case prior gives 0 for the posterior confidence in the bound $\mathit{pfm}_B<p_B$, and thus 0 as the infimum for \eqref{eq_post_conf_2D}. Consequently, we only (non-trivially) consider the next case. 
			\item when $\phi>1-\theta$, to minimise the objective function $h$, we set $M_5=\phi-1+\theta$, $M_1=1-\phi$, $M_3=1-\theta$ and $M_2=M_4=M_6=M_7=0$. That corresponds to the worst-case 3-point prior depicted in Fig.~\ref{fig_worst_case_joint_prior}, which is $Pr(X=x,Y=y)=(1-\phi){\bf 1}_{x=p_l,y=p_B} + (1-\theta){\bf 1}_{x=p_B,y=p_B}+(\phi-1+\theta){\bf 1}_{x=\epsilon,y=\epsilon}$. Using this prior distribution, the value of $Pr(Y < p_B \mid n_A, n_B)$ is given by \eqref{eq_worst_case_2d_posterior}. All of the preceding arguments guarantee that this value must be the infimum for the posterior confidence $Pr(Y \leqslant p_B \mid n_A, n_B)$, as claimed in Theorem \ref{theorem_two_d_CBI}.
		\end{itemize}			
	\end{proof}

	Moreover, to properly answer questions Q4 and Q5, we need to assign a required level of confidence $c$ to \eqref{eq_worst_case_2d_posterior}, and then solve it for $n_B$:
	
	\begin{equation}
	n_B=\frac {
	\ln 
		\left( {\frac { \left(  \left( 1-\theta \right)  \left( 1-{\it p_B}
				\right) ^{{\it n_A}}+ \left( 1-{\it p_l} \right) ^{{\it n_A}} \left( 
				1-\phi \right)  \right) c}{ \left( \phi-1+\theta \right)  \left( 1-c
				\right) }} \right)
		-\ln  \left( 1-\epsilon \right)^{n_A}
	}
	{\ln  \left( 1-\epsilon
		\right)-\ln  \left( 1-{\it p_B} \right)  }
	\label{eq_app_nB}
	\end{equation}
	
	The result \eqref{eq_app_nB} is also used to generate Fig.~\ref{fig_nA_vs_nB_two_d} by fixing $c$, $p_B$, $p_l$, $\epsilon$, $\phi$, $\theta$ and treating $n_A$ as an independent variable. By a monotonicity analyses -- taking partial derivatives $\frac{\partial n_B(n_A)}{\partial n_A}$ and solving for $n_A$ -- we find that the minimum point on the curves in Fig.~\ref{fig_nA_vs_nB_two_d} is located at
	\begin{equation}
	n_A=\frac {\ln  \left( -{\frac {1-\theta}{1-\phi}\ln  \left( {\frac {1-{\it p_B}}{1-\epsilon}} \right)  \left( \ln  \left( {\frac {1-{\it p_l}}{1-\epsilon}} \right)  \right) ^{-1}} \right)}{\ln  \left( 1-{\it p_l} \right) -\ln  \left( 1-{\it p_B}	\right) }
	\label{eq_app_nA}
	\end{equation}
	
	\section{$n_B$ is unbounded as $n_A$ grows}
	\label{sec_app_D}
	Recall that $p_l < \epsilon < p_B$. Upon rewriting \eqref{eq_app_nB}, we have
	
	\begin{align}
	n_B=\frac {
	\ln \left( \frac{ 1-{p_l}}{1-\epsilon}
				\right)^{n_A} + \ln 
		\left( { \frac { \left(  \left( 1-\theta \right)\left(\frac{ 1-{p_B} }{ 1-{p_l}
				}\right) ^{n_A}  +  \left( 
				1-\phi \right)  \right) c}{ \left( \phi-1+\theta \right)  \left( 1-c\right) }} \right)
	}
	{\ln  \left( 1-\epsilon
		\right)-\ln  \left( 1-{\it p_B} \right)  }
	\label{nBgrowswithnA}
	\end{align}
	The inequalities above imply both $\left(\frac{ 1-{p_l}}{1-\epsilon}\right)>1$
	and $\left(\frac{ 1-{p_B} }{1-{p_l}}\right)<1$.
Therefore, $n_B \to \infty$ as $n_A \to \infty$, and the growth of $n_B$ is $O(n_A)$.

	\section*{Acknowledgements}
	This work was supported in part by the Intel Collaborative Research Institute on Safe Automated Vehicles (ICRI-SAVe), by the UK EPSRC through the Offshore Robotics for Certification of Assets (ORCA) [EP/R026173/1], and ORCA's Partnership Resource Fund through Continual Verification and Assurance of Robotic Systems under Uncertainty (COVE). We thank Bev Littlewood for insightful comments on earlier versions of the paper; and
	the anonymous reviewers who gave very helpful comments for the final version of the paper.
	

	
	\bibliography{ref}

\end{document}